\documentclass[
twoside
]{article}
\usepackage{arxvis,palatino,epsfig,latexsym,natbib}

\usepackage{caption}
\usepackage{subcaption}
\usepackage{hyperref}
\usepackage{multirow}
\usepackage{booktabs}
\usepackage{tabularx}
\usepackage{makecell}
\usepackage{amsmath}
\usepackage{wasysym}
\usepackage{xspace}
\usepackage{amssymb}
\usepackage{soul}

\usepackage{lineno}
\usepackage{xcolor}

\usepackage{tikz}
\usetikzlibrary{angles, quotes}
\usepackage{pgfplots}

\usetikzlibrary{shapes, arrows, arrows.meta, fit, calc, patterns, positioning, decorations, backgrounds, chains,fit, shadings}
\usetikzlibrary{plotmarks}
\tikzset{>=latex}
\pgfplotsset{compat=1.18}

\definecolor{CBblue}{RGB}{27, 158, 119}%{0, 73, 73} % Protanopia-friendly blue
\definecolor{CBorange}{RGB}{217, 95, 2}%{219, 132, 61} % Deuteranopia-friendly orange
\definecolor{CBgreen}{RGB}{117, 112, 179}%{0, 128, 0} % Tritanopia-friendly green

\newcommand{\R}{\mathbb{R}}

\newcommand{\N}{\mathbb{N}}
\newcommand{\ND}{\text{ND}}
\newcommand{\dom}{\preceq}

\newcommand{\decSample}{X}
\newcommand{\decSampleSize}{N}
\newcommand{\decSpace}{\mathcal{X}}
\newcommand{\decSpaceDim}{d}
\newcommand{\objSpace}{\mathcal{Y}}
\newcommand{\objSpaceDim}{m}
\newcommand{\ndsLayer}{L}
\newcommand{\elaSample}{S}

\newcommand{\gDec}{G_{D}}

\newcommand{\stGraph}{G^{\text{\textnormal{ST}}}}
\newcommand{\stObj}{G^{\text{\textnormal{ST}}}_{O}}
\newcommand{\stObjToDec}{G^{\text{\textnormal{ST}}}_{O \to D}}
\newcommand{\stDec}{G^{\text{\textnormal{ST}}}_D}
\newcommand{\stDecToObj}{G^{\text{\textnormal{ST}}}_{D \to O}}

\newcommand{\knnGraph}{G^{k\text{\textnormal{NN}}}}
\newcommand{\knnObj}{G^{k\text{\textnormal{NN}}}_{O}}
\newcommand{\knnObjToDec}{G^{k\text{\textnormal{NN}}}_{O \to D}}
\newcommand{\knnDec}{G^{k\text{\textnormal{NN}}}_D}
\newcommand{\knnDecToObj}{G^{k\text{\textnormal{NN}}}_{D \to O}}

\newcommand{\MOEAD}{MOEA/D\xspace}
\newcommand{\NSGAII}{NSGA-II\xspace}
\newcommand{\SMSEMOA}{SMS-EMOA\xspace}

\begin{document}

\title{\bf MO-ELA: Rigorously Expanding Exploratory Landscape Features for Automated Algorithm Selection in Continuous Multi-Objective Optimisation}  

\author{\bf Oliver Ludger Preu\ss \hfill oliver.preuss@upb.de\\ 
        Machine Learning and Optimisation, Paderborn University, Germany
\AND
        \bf Jeroen Rook \hfill jeroen.rook@upb.de\\
        Machine Learning and Optimisation, Paderborn University, Germany
\AND
        \bf Jakob Bossek \hfill jakob.bossek@upb.de\\
        Machine Learning and Optimisation, Paderborn University, Germany
\AND
        \bf Heike Trautmann \hfill heike.trautmann@upb.de\\
        Machine Learning and Optimisation, Paderborn University, Germany\\
        Data Management \& Biometrics, University of Twente, The Netherlands
}

\maketitle

\begin{abstract}

Automated Algorithm Selection (AAS) is a popular meta-algorithmic approach and has demonstrated to work well for single-objective optimisation in combination with exploratory landscape features (ELA), i.e., (numerical) descriptive features derived from sampling the black-box (continuous) optimisation problem. In contrast to the abundance of features that describe single-objective optimisation problems, only a few features have been proposed for multi-objective optimisation so far.
Building upon recent work on exploratory landscape features for box-constrained continuous multi-objective optimization problems, we propose a novel and complementary set of additional features (MO-ELA). 
These features are based on a random sample of points considering both the decision and objective space. The features are divided into $5$ feature groups depending on how they are being calculated: \textit{non-dominated-sorting, descriptive statistics, principal component analysis, graph structures and gradient information.}
An AAS study conducted on well-established multi-objective benchmarks demonstrates that the proposed features contribute to successfully distinguishing between algorithm performance and thus adequately capture problem hardness resulting in models that come very close to the virtual best solver. After feature selection, the newly proposed features are frequently among the top contributors, underscoring their value in algorithm selection and problem characterisation.

\end{abstract}


\section{Introduction}
\label{sec:introduction}

In (heuristic) optimisation there is rarely an algorithm that performs best on any problem instance. Instead, usually a set of algorithms exist that show complementary performance. A well established and sophisticated approach termed \emph{Automated Algorithm Selection}~(AAS)~\cite{KerEtAl2019} leverages these complementarities by constructing machine learning models that predict the likely best-performing algorithm on a per-instance basis.  First proposed by Rice~\cite{Rice1976} decades ago AAS is now a mature field of research. AAS relies on characteristics of optimisation problems termed \emph{instance features} or plain \emph{features}. Features describe the properties of problem instances, like multi-modality and deceptiveness, and can ideally be calculated with affordable computational effort prior to an algorithm run. 

Algorithm selection techniques have achieved considerable successes in addressing a wide range of computationally demanding problem domains. These include Constraint Programming~\cite{omahony_using_2008}, propositional satisfiability problem (SAT)~\cite{xu_satzilla_2008,xu_evaluating_2012}, Mixed-Integer Programming~(MIP)~\cite{xu_hydra-mip_2011}, AI planning \cite{seipp_learning_2012}, and the Travelling Salesperson Problem~(TSP)~\cite{KerErAl2018,PihMus2014} to name a few; see, e.g., \cite{KerEtAl2019} for a recent survey with more examples.

Moreover, AAS has been successfully applied in the field of black-box continuous single-objective (SO) optimisation~\cite{MerEtAl2011,BisEtAl2012} with a plethora of introduced features. Recently, also feature-free approaches have been investigated based on deep-learning. \cite{SeilerEtAl2020} studied such an approach in the context of algorithm selection for the TSP while in a later work a collection of deep learning-based feature-free approaches for characterising single-objective continuous fitness landscapes was proposed~\cite{SeilerEtAl2022}. Alternatively, \citet{van2023doe2vec} proposed DoE2Vec which is an autoencoder that learns optimisation landscapes characteristics which can be used for AAS. \cite{cenikj2024transoptas} proposed a transformer-based algorithm selector which directly predicts algorithm performance. However, all of the mentioned deep learning approaches focus on SO which would need adaption to work on MO problems.

However, features in the domain of continuous multi-objective~(MO) optimisation problems are rare. A straight-forward approach is to calculate single-objective features for each objective function individually~\cite{KerTra2016}. While this is a legitimate approach and these features apparently capture the individual function properties, they fail to capture complex interactions between the objective functions. Only recently research in this field gained momentum. In 2020 \cite{LieEtAl2020} introduced combinatorial multi-objective ELA (MO-ELA) features based on random and adaptive walks. A year later the same authors adapted their set of combinatorial MO-ELA features to continuous MO-problems (MOP)~\cite{LieEtAl2021}. \cite{PicSch2021} introduced so-called MO-violation landscapes where violation of said constraint are integrated into the fitness function. Their work was later extended by introducing further features by \cite{VodEtAl2022} and \cite{AlsEtAl2023}.
Most recently \cite{seiler2025deep} developed a deep learning transformer based architecture that can be used for MO continuous problems.

In this paper, building upon the recent work of Liefooghe \emph{et al.}~\cite{LieEtAl2021}, we introduce a new set of ELA features tailored for box-constrained continuous multi-objective optimisation problems. These features are primarily based on dominance relationships within a sample of solutions and leverage various graph-based representations in both decision and objective spaces. An AAS study on standard multi-objective benchmark problems demonstrates that these features significantly improve the performance of selection models. Following feature selection, they consistently rank among the most influential features, highlighting their effectiveness in guiding algorithm selection for multi-objective optimisation tasks.

The paper is structured as follows: In Section~\ref{sec:preliminaries} we introduce multi-objective optimisation and revisit existing MO-ELA features. Next, in Section~\ref{sec:features} we introduce our novel set of features emphasise similarities to existing features. Section~\ref{sec:experiments} is dedicated to a comprehensive AAS study that showcases the usefulness of our features followed by a discussion in Section~\ref{sec:discussion}. We finalise the paper with concluding remarks and an outlook on future work in Section~\ref{sec:conclusion}.

\section{Preliminaries}
\label{sec:preliminaries}

In this section we first introduce multi-objective optimisation problems and later discuss MO-ELA features from the literature.

\subsection{Multi-Objective Optimisation}

The majority of real-world problems is composed of multiple, say $\objSpaceDim \in \N_{+}$, objective functions $f_i : \decSpace \to \R$ for $i \in [\objSpaceDim]$ where $[\objSpaceDim] := \{1, \ldots, \objSpaceDim\}$. Hence, we are faced with a \emph{multi-objective optimisation problem}~(MOP)
\begin{align*}
  f : \decSpace \to \objSpace \subseteq \R^{\objSpaceDim}, x \mapsto \left(f_1(x), \ldots, f_{\objSpaceDim}(x)\right)^{\top}
\end{align*}
where the objective function is vector-valued. $\mathcal{X}$ is the problem-specific decision space. In our setting $\decSpace = \mathbb{R}^{\decSpaceDim}$ for some positive integer $d$, i.e., we are dealing with continuos optimisation. Objectives are usually conflicting, meaning that decreasing one objective is only possible while increasing at least one other objective. 
In order to do optimisation in this context we need a preference relation $x \dom y$ on $\decSpace$ which indicates when solution $x$ is better than solution $y$.
In this paper, we adopt the well-known \emph{Pareto-dominance} relation~(see, e.g.,~\cite{Ehrgott2005}). 
For $x, y \in \decSpace$ we write $x \dom y$ and say \emph{$x$ dominates $y$} and likewise $y$ \emph{is dominated} by $x$, if $f_i(x) \leq f_i(y)$ for 
all $i \in [\objSpaceDim]$ and $\exists j \in [\objSpaceDim]$ such that $f_j(x) < f_j(y)$. 
Informally, $x$ dominates $y$, if $f(x)$ is not worse than $f(y)$ in all objectives and it is strictly better in at least one objective.
Let $X \subseteq \decSpace$ be a set of points. We denote by $\text{ND}(X) = \{x \in X \mid \nexists y \in X: f(y) \dom f(x)\}$ the subset of \emph{non-dominated} points in set $X$. 
The goal in MOO is to calculate the \emph{Pareto-set} $\decSpace^{*} = \text{ND}(\decSpace)$ and its image under $f$, the \emph{Pareto-front} $f(\decSpace^{*}) = \{f(x) \mid x \in \decSpace^{*}\}$.
Points in $\decSpace^{*}$ and $f(\decSpace^{*})$ are termed \emph{Pareto-optimal}.

\subsection{Existing Features}

\begin{table}
    \caption{Overview of papers that deal with (MO)-ELA}
    \footnotesize
    \renewcommand{\arraystretch}{1.6}
    \centering
    \begin{tabularx}{\columnwidth}{XlX}
        \toprule
        {\bf Author(s)} & {\bf Domain} & {\bf Comment} \\
        \midrule
        \cite{LieEtAl2020} & comb. MOP & MO-ELA features for combinatorial optimisation problems based on random and adaptive walks.\\
        \cite{LieEtAl2021} & interpolated cont. MOP & Adapted the combinatorial features of \cite{LieEtAl2020} to the continuous MO domain \\
        \cite{PicSch2021} & CMOP & Introduced MO violation landscapes and based some features on that. \\
        \cite{VodEtAl2022} & CMOP & Used MO violation landscapes of Picard \& Schiffmann '21 and extended it with more features  \\
        \cite{AlsEtAl2023} & CMOP & Used MO violation landscapes of \cite{PicSch2021} and the features of \cite{VodEtAl2022} and extended it further.\\
        \cite{KerTra2016} & SOP $\rightarrow$ MOP & Application of SO features on MOPS \\
        \cite{seiler2025deep} & SOP / MOP &  Self-Supervised Pretrained Transformers for Single- and Multiobjective Continuous Optimization Problems \\
        
        \bottomrule
    \end{tabularx}
    \label{tab:feature_overview_compact}
\end{table}

Table \ref{tab:feature_overview_compact} contains an -- not exhaustive -- overview of existing literature regarding MO-ELA. 

\citet{KerTra2016} applied single-objective ELA features to multi-objective problems (MOP). They calculated single-objective ELA features for each of the objectives of bi-objective problems and used the ratio of the resulting single-objective feature vectors as a feature vector for the MOP. The results were indicative as the complex structure of MOPs and interdependencies of objectives were not captured. Leading to the conclusion that tailored MO-ELA features are needed.

Three papers \cite{PicSch2021, VodEtAl2022, AlsEtAl2023} describe how to characterise constrained multi-objective problems (CMOP). These MO-ELA features are mostly based on violation landscapes. Violation landscapes are calculated by replacing the fitness function with a constraint violation function. Since we will use multi-objective continuous problems without constraints -- except box constraints -- these feature sets are not directly applicable.

\cite{LieEtAl2020} introduced MO-ELA features for combinatorial MOPS. These features are based on full enumerations and random and adaptive walks of the problem.
Later \cite{LieEtAl2021} adapted the MO-ELA features for combinatorial MOPS for (interpolated) continuous optimisation problems. The features are categorized into four groups -- global, multimodality, evolvability, and ruggedness. Each group aims to capture specific properties of a problem. This is done by calculating a random sample of the problem, e.g. using Latin hypercube sampling (LHS). For each point in this sample, the closest $\decSpaceDim$ neighbours are determined based on the Euclidean distance in the decision space. The sample in combination with the defined neighbourhood is used to calculate the features. 
\textit{Multimodality} describes the presence of local and global optima. To capture this the number of single and multi-objective local optima in the sample and multiple adaptive walks based on the neighbourhood are performed that capture the size and amount of basins present in the problem. \textit{Evolvability} captures the possible improvement through the neighbourhood of a solution. Here features like the number of dominating, dominated or incomparable neighbours are calculated or the distance to the neighbours on decision and objective space as well as the ratio of these distances. \textit{Ruggedness} describes the correlation of information among neighbouring solutions. This is calculated by taking the ratio of the features calculated by evolvability. A higher correlation indicates a smoother landscape. Combining all features a total of $53$ and $59$ features for bi- and tri-objective problems respectively are created. In the performed experiments of this paper, these features are used in addition to the newly proposed ones.

\subsection{Feature-Free Approaches}

In contrast to handcrafted features, recent work~\citep{SeilerEtAl2022,van2023doe2vec,cenikj2024transoptas} proposed deep learning–based approaches for single-objective optimization (SO), in which numerical descriptive features are not manually constructed but instead learned automatically using deep learning techniques. These methods are therefore referred to as feature-free approaches, as handcrafted features are no longer required. Their main advantage lies in the automated feature learning process, while their main drawback is the lack of interpretability, since the meaning of individual learned features is generally not accessible to the user.

Beyond interpretability, the two paradigms also differ in their data and computational requirements. In particular, ELA-based methods do not need any training instances compared to deep learning–based feature-free approaches, which is especially beneficial when optimization problems are scarce or when the generation of training data is computationally expensive. By contrast, deep learning models may still generalize poorly in settings with limited training data, even when data augmentation techniques are applied. On the other hand, feature free methods are fast to compute and are generally faster than handcrafted features.

A naive way to adapt the feature free approach to multi-objective problems is to calculate the single-objective representation of the individual objectives and combine them into a single vector using the ratio of the single vectors as we described earlier where \cite{KerTra2016} applied SO-ELA to MOPs. 
Only recently, \cite{seiler2025deep} proposed a self-supervised pretrained transformer that can be used to latently characterise a fitness landscape of SO and MO continuous optimisation problems directly. For this paper we decided not include this technique as it is constrained to have $d+m \leq 12$. This means that for bi-objective functions we can have a $d$ of at most $10$ and for tri-objective functions we can have a $d$ of at most $9$. Additionally, the sample size has to be either $25d$ or $50d$. These constrains are, in our view, too limiting to have a meaningful comparison with these feature-free features. However, we plan to investigate the potential of these features in future studies.

\section{Features}
\label{sec:features}

\begin{figure}
    \centering
    \begin{tikzpicture}[scale=0.55]

    % ===
    % DECISION SPACE
    % ===

    \begin{scope}[scale=0.5, every node/.style={font=\footnotesize}] % No transformation

        % axis
        \draw[->] (0,0) to (14, 0);
        \draw[->] (0,0) to (0, 14);

        % axis labels
        \node[rotate=90] at (-0.75, 7) {\footnotesize{$x_1$}};
        \node at (7, -0.75) {\footnotesize{$x_2$}};

        % grid
        \foreach \i in {1, ..., 13} {
            \draw[gray!30] (\i, 13) -- (\i, 0);% node[below, font=\footnotesize] {\textcolor{black}{$\i$}};
            \draw[gray!30] (13, \i) -- (0, \i);% node[left, font=\footnotesize] {\textcolor{black}{$\i$}};
        };

        % decision space label
        \node (X) at (6.5,14) {Decision space $\decSpace \subseteq \R^2$};

        % coordinate for drawing an arrow from decision to objective space
        \coordinate (dec) at (11, 8);

        % % examplary decision space points (later mapped to objective space points)
        % \foreach lab/na/x/y in {x_1/x1/8/4, x_2/x2/7/10, x_3/x3/9/6} {
        %   \coordinate[label=below:\scriptsize{$\lab$}] (\na) at (\x, \y);
        %   \filldraw[black] (\na) circle (5pt);
        % }

        % place point coordinates
        \coordinate[label=below:\scriptsize{$x_1$}] (x1) at (8,4);
        \coordinate[label=left:\scriptsize{$x_2$}] (x2) at (7,10);
        \coordinate[label=left:\scriptsize{$x_3$}] (x3) at (9,6);
        \coordinate[label=left:\scriptsize{$x_4$}] (x4) at (2,3);
        \coordinate[label=left:\scriptsize{$x_5$}] (x5) at (2,7);
        \coordinate[label=left:\scriptsize{$x_6$}] (x6) at (12,12);

        % draw MST edges
        \draw (x5) edge[-] (x4);
        \draw (x4) edge[-] (x1);
        \draw pic["$\alpha_x$", draw=black, angle eccentricity=0.55, angle radius=0.4cm]{angle = x1--x4--x5};
        
        % draw points
        \filldraw[CBblue] (x1) circle (5pt);
        \filldraw[CBgreen] (x2) circle (5pt);
        \filldraw[CBgreen] (x3) circle (5pt);
        \filldraw[CBblue] (x4) circle (5pt);
        \filldraw[CBblue] (x5) circle (5pt);
        \filldraw[CBorange] (x6) circle (5pt);

    \end{scope}

    % ===
    % OBJECTIVE SPACE
    % ===

    % Second coordinate system
    \begin{scope}[shift={(8,0)}, scale=0.5, every node/.style={font=\footnotesize}]%, rotate=45, scale=0.5] % Shifted, rotated, and scaled coordinate system

        % axis
        \draw[->] (0,0) to (14, 0);
        \draw[->] (0,0) to (0, 14);

        % axis labels
        \node[rotate=90] at (-0.75, 7) {\footnotesize{$f_2 \to \min!$}};
        \node at (7, -0.75) {\footnotesize{$f_1 \to \min!$}};

        % grid
        \foreach \i in {1, ..., 13} {
            \draw[gray!30] (\i, 13) -- (\i, 0);% node[below, font=\footnotesize] {\textcolor{black}{$\i$}};
            \draw[gray!30] (13, \i) -- (0, \i);% node[left, font=\footnotesize] {\textcolor{black}{$\i$}};
        };

        % points

        % objective space label
        \node (Y) at (6.5,14) {Objective space $\objSpace \subseteq \R^2$};

        % coordinate for drawing an arrow from decision to objective space
        \coordinate (obj) at (3, 8);

        % place point coordinates (in objective space)
        \coordinate[label=below:\scriptsize{$y_1$}] (fx1) at (4,2.65);
        \coordinate[label=right:\scriptsize{$y_2$}] (fx2) at (6,9);
        \coordinate[label=right:\scriptsize{$y_3$}] (fx3) at (9,5);
        \coordinate[label=right:\scriptsize{$y_4$}] (fx4) at (8,1);
        \coordinate[label=right:\scriptsize{$y_5$}] (fx5) at (2,6);
        \coordinate[label=right:\scriptsize{$y_6$}] (fx6) at (12,10.5);

        % draw MST edges
        \draw (fx5) edge[-] (fx4);
        \draw (fx4) edge[-] (fx1);
        \draw pic["$\alpha_y$", draw=black, angle eccentricity=1.5, angle radius=0.45cm]{angle = fx5--fx4--fx1};
        
        % draw points
        \filldraw[CBblue] (fx1) circle (5pt);
        \filldraw[CBgreen] (fx2) circle (5pt);
        \filldraw[CBgreen] (fx3) circle (5pt);
        \filldraw[CBblue] (fx4) circle (5pt);
        \filldraw[CBblue] (fx5) circle (5pt);
        \filldraw[CBorange] (fx6) circle (5pt);

    \end{scope}

    %\draw (dec) edge[-latex, thick, bend left=25] node[above] {$f : \mathcal{X} \to F$} (obj);
    
    % Arrows between points in decision space and their objective vectors
    \draw (x1)++(0.1,0) edge[-latex, thick, gray, dashed, bend right = 5] (fx1);
    \draw (x2)++(0.1,0) edge[-latex, thick, gray, dashed, bend left = 10] (fx2);
    \draw (x3)++(0.1,0) edge[-latex, thick, gray, dashed, bend right = 11] (fx3);
    \draw (x4)++(0.1,0) edge[-latex, thick, gray, dashed, bend right = 7] (fx4);
    \draw (x5)++(0.1,0) edge[-latex, thick, gray, dashed, bend left = 14] (fx5);
    \draw (x6)++(0.1,0) edge[-latex, thick, gray, dashed, bend left = 10] (fx6);

\end{tikzpicture}
    \caption{Visualisation of the MO-ELA approach for a function $f : \decSpace \subseteq \R^2 \to \R^2$. The decision space with six sampled points $x_1, \ldots, x_6$ is shown to the left while the respective objective space is depicted to the right. Points are coloured by non-domination layers. Here, \textcolor{CBblue}{$\ndsLayer_1 = \{(x_1, y_1), (x_4, y_4), (x_5, y_5)\}$}, \textcolor{CBgreen}{$\ndsLayer_2 = \{(x_2, y_2), (x_3, y_3)\}$} and \textcolor{CBorange}{$\ndsLayer_3 = \{(x_6, y_6)\}$}. Edges in the left plot correspond to the edges of the MST $\stDec$ while the edges in the right plot illustrate the respective spanning tree $\stDecToObj$. 
    }
    \label{fig:mo-ela-visualisation}
\end{figure}
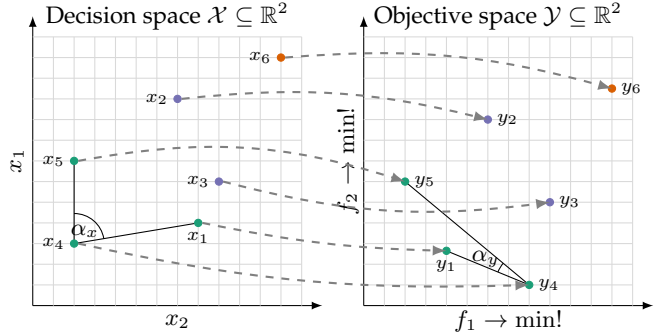

Here we will provide detailed descriptions of the newly proposed features.
These features are divided into 5 groups with a total of 226 and 233 features for 2 and 3 objectives respectively. The objectives are min-max scaled independently based on the random sample and the decision space is scaled into $[0, 1]$ based on the given box-constrains to make features comparable even for arbitrary ranges across different functions before feature calculation. 

In the following we introduce some vocabulary as a foundation for a solid mathematical description of the introduced features. 
We sample a random set $\decSample = \{x_1, \ldots, x_N\} \subseteq \decSpace$ in the decision space of cardinality $\decSampleSize$. Let $y_i = f(x_i) \in \R^{\objSpaceDim} \, \forall i \in [\decSampleSize]$ be the respective objective function values. This yields
\begin{align*}
    \elaSample = \{s_1 = (x_1, y_1), \ldots, s_{\decSampleSize} = (x_{\decSampleSize}, y_{\decSampleSize})\}
\end{align*}
that serves as a foundation for the feature calculation (see Figure~\ref{fig:mo-ela-visualisation} for a visualisation). For ease of readability we overload 
the adopted preference relation (see Section~\ref{sec:preliminaries}) and write $s_i \dom s_j$ if and only if $y_i \dom y_j$. This allows us to conveniently write $\ND(\elaSample)$ to obtain the non-dominated sub-set of $\elaSample$.

\subsection{Non-Dominated Sorting Based Features}

The first group of features is based on non-dominated sorting~(NDS) of $\elaSample$. NDS was first introduced by Goldberg~\cite{Goldberg1989} and later adopted successfully as the foundation for evolutionary multi-objective algorithms, predominately the non-dominated sorting genetic algorithm~(\NSGAII)~\cite{DebEtAl2002}. The procedure sorts $\elaSample$ into $h \in [N]$ disjoint layers $\ndsLayer_1, \ndsLayer_2, \ldots, \ndsLayer_h$ where the first layer only contains non-dominated points, i.e., $\ndsLayer_1 = \text{ND}(\elaSample)$. Consecutive layers $L_i, i > 1$ are constructed as the set of non-dominated points ignoring all points from previous layers, i.e., $\ndsLayer_i = \text{ND}(\elaSample \setminus (\ndsLayer_1 \cup \ndsLayer_2 \cup \ldots \cup \ndsLayer_{i-1}))$.

\begin{figure}
    \centering
    \includegraphics[width=0.4\linewidth]{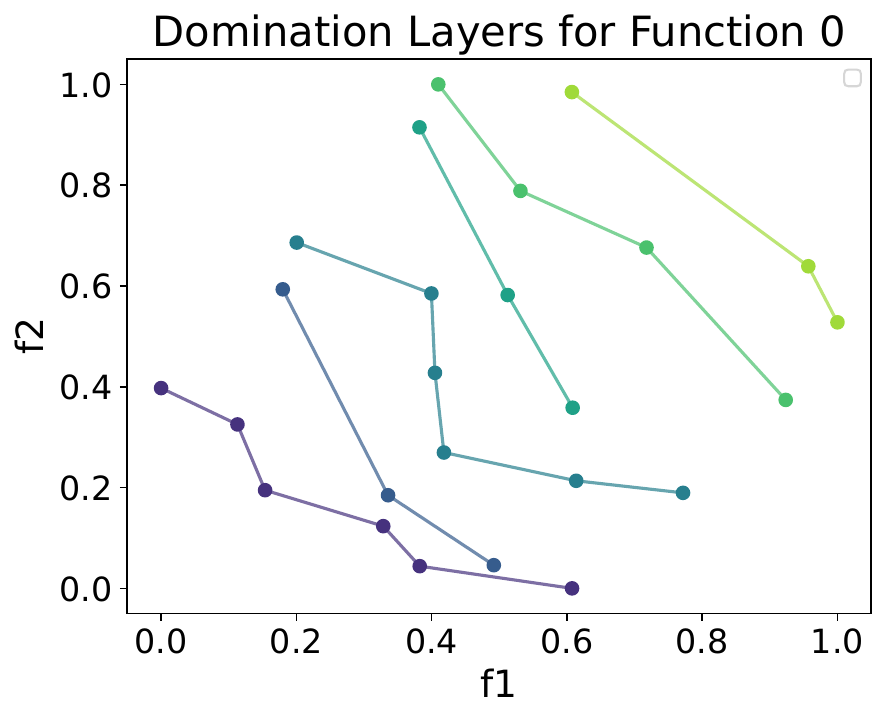}
    \includegraphics[width=0.4\linewidth]{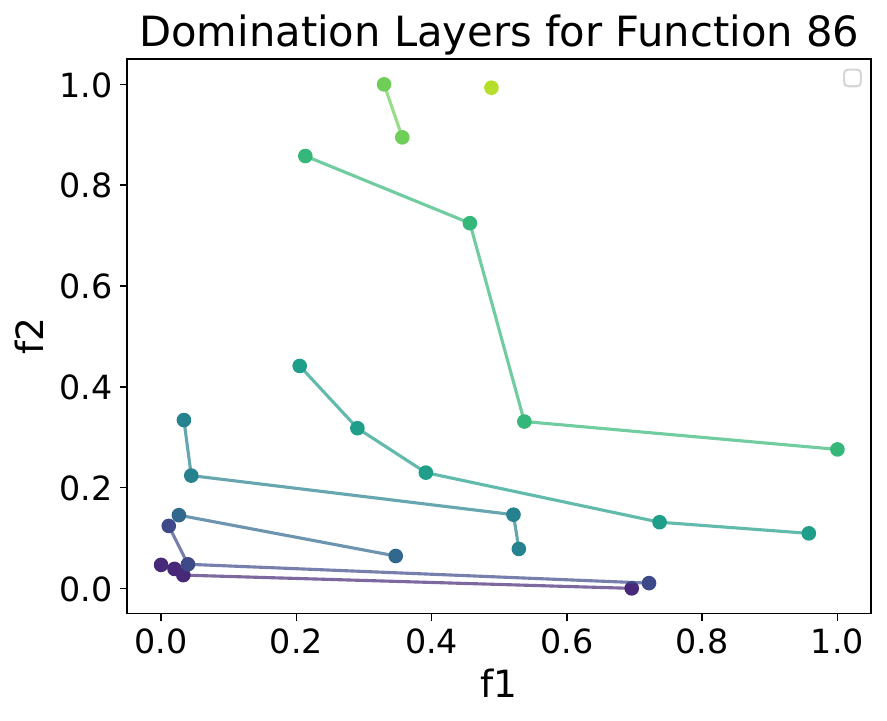}
    
    \includegraphics[width=0.4\linewidth]{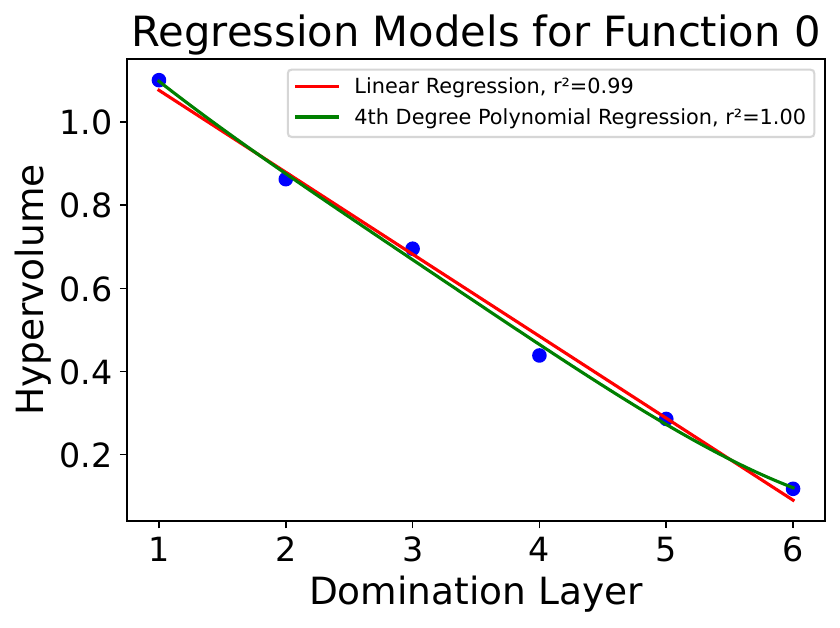}
    \includegraphics[width=0.4\linewidth]{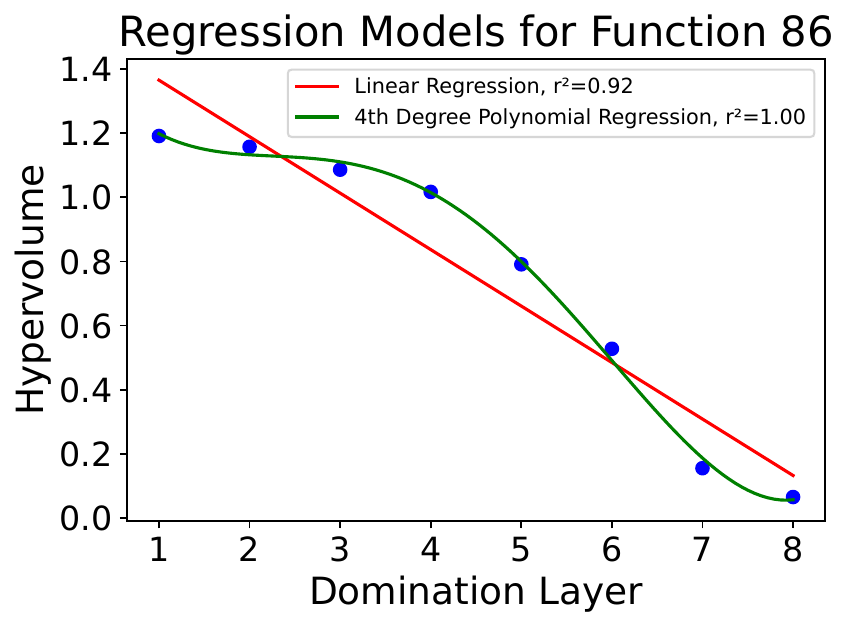}
    \caption{Domination layers for BiObjBBOB functions $1$ and $86$ and polynomial regression of the corresponding HV values with degrees $1$ and $4$.}
    \label{fig:hv_regression}
\end{figure}

Based on this partition the number of non-dominated points $|L_1|$ (\texttt{no\_non\_dom\_points}), the number of layers $h$ (\texttt{max\_rank}), the average number of points per layer (\texttt{avg\_points\_per\_layer}) are considered as features. Additionally, performance metrics per layer $\ndsLayer_i, i \in [h]$ can be calculated. We chose the first~5 layers and calculate the Hypervolume~(HV)~\cite{ZitzlerTLFF03} (\texttt{hv\_dom\_layer\_$i$}) and Solow-Polasky measure~(SP) \cite{SP2006diversity} (\texttt{sp\_dom\_layer\_$i$}) per layer. The reference point for the HV has been set to $[1.1]^\objSpaceDim$, since the objectives are being scaled to $[0,1]$. We add the offset of $0.1$ to the reference point to allow extreme points to have a contribution.
In case there are less than $5$ layers HV and SP are set to $0$ for the missing layers. 
For the last features, a (non-)linear regression with polynomial degrees $p \in \{1, 2, 3, 4\}$ is used with the HV of all layers as target input. The features are the resulting $R^2$ indicators (\texttt{r$p$}). These features indicate the overall structure of the 
problem. If the problem is easy, e.g., a simple sphere -- BiObjBBOB function $1$ -- all sampled points will be distributed equally over all domination layers with all layers having an approximately equal distance resulting in a linear decreasing HV per layer. When performing a regression the $R^2$ of $p = 1$ and $p=4$ will both be high since both functions can fit a line (Fig. \ref{fig:hv_regression} left column). For hard problems like a highly multi-modal function -- e.g. BiObjBBOB function $86$ -- the sampled points will be distributed unequally into the different layers and the layers have different distances to each other depending on the local optima present in the problem resulting in a non-linear decreasing HV per layer. The resulting regression will have a low $R^2$ with $p = 1$ and a high $R^2$ with $p = 4$, since the HV distribution forms a curve which the non-linear regression with $p = 4$ can fit (Fig. \ref{fig:hv_regression} right column). In total the NDS feature group consists of $19$ features.

\subsection{Statistics of Non-Dominated Points}

The next feature group are simple descriptive statistics only applied to the objective vectors of non-dominated points $Y_1 = \{y \mid (x, y) \in L_1\}$. Here, the minimal, maximal, and average value per objective are used alongside the respective standard deviation (\texttt{min|max|std|avg}), as well as the pairwise (Spearman) correlation of the objectives (\texttt{(spearman\_)corr\_obj}). 
For $\objSpaceDim = 2$ the absolute difference of the standard deviation of both objectives is used (\texttt{obj\_std\_diff}). If $|\ndsLayer_1| = 1$ the (Spearman) correlation is set to~$1$. In total this creates $11$ ($12$) features for $2$ ($3$) objectives respectively.

\subsection{PCA-Based Features}

This feature group is based on a \emph{principal component analysis}~(PCA, \cite{Pearson1901}). Like the previous feature groups this group only takes the non-dominated points $\ndsLayer_1$ into account. 
The PCA is applied to three sets; the observations $X = \{x \mid (x, y) \in \ndsLayer_1\}$ in decision space, the respective objective space vectors $Y = \{y \mid (x, y) \in \ndsLayer_1\}$ and the concatenated full design $X\_Y =\left\{x\frown y \mid (x, y) \in \ndsLayer_1 \right\}$.
For each of the three analyses each the principal components with the lowest, highest, and average explained variance of all components (\texttt{($X$|$Y$|$X\_Y$)(min|max|avg)}) are used as features. In total this creates 9 features. 

\subsection{Graph-Based Features}

This sub-set of features is based on different graphs derived from non-domination layer $L_1$. Hence, we first describe the construction of these graphs.

We first build a complete undirected, weighted graph $\gDec = (V, E)$ (the sub-script $D$ indicates the decision space; likewise, later, the sub-script $O$ will indicate the objective space and $O \to D$ or $D \to O$ respectively will indicate a \emph{transfer}). Here, $V = [\decSampleSize]$, i.e., the indices of our sample $\elaSample$. The edge set $E$ contains all $\binom{\decSampleSize}{2}$ pairwise edges $\{i, j\}, 1 \leq i \neq j \leq \decSampleSize$ with the respective weight set to $w(i, j) = \text{dist}(x_i, x_j)$ where $\text{dist} : \decSpace \times \decSpace \to \R_{+}$ is the Euclidean distance\footnote{We highlight though, that \underline{any} other meaningful distance metric is possible; even distances in combinatorial domains like the Hamming distance.} in the decision space. Next, we calculate a minimum spanning tree~(MST; see, e.g., \cite{CorEtAl2022_algorithms})\footnote{Any connected, acyclic sub-graph of a graph is a spanning tree. A minimum spanning tree is a spanning tree of minimum total weight in a weighted graph.} 
$\stDec = (V, E')$ of $\gDec$~(see solid edges in left plot in Figure~\ref{fig:mo-ela-visualisation} for an example). Using $\stDec$ we build another spanning tree $\stDecToObj$ with weights set to $w(i, j) = \text{dist}(y_i, y_j)$, i.e., the (Euclidean) distances in objective space. 
I.e., we \underline{transfer} the MST-edges from the decision space graph to the objective space graph (termed \emph{edge transfer}). 
Note that $\stDecToObj$ is guaranteed to be a spanning tree, but not necessarily a minimum spanning tree in objective space as exemplified in Figure~\ref{fig:mo-ela-visualisation}~(right plot).
This construction is also done the other way around: we first build an MST on the non-dominated points in objective space and do an edge transfer to the points in decision space yielding $\stObj$ and $\stObjToDec$. 

Analogously, using $k$-nearest-neighbour graphs~($k$NN) instead of MSTs with $k=1$, we obtain additional four graphs $\knnDec$, $\knnDecToObj$, $\knnObj$, and $\knnObjToDec$.

The rationale for creating a graph for the decision and objective space, and transferring the nodes is as follows. The shape of the points for the decision and objective space can take very different forms depending on the problem.
For example, the decision space of a problem can have multiple clusters which create multiple connected components in the $\knnDec$ while the objective space forms a convex front which creates one connected component in the $\knnObj$. This difference already creates different features specific to the problem. When transferring the edges in a next step the $\knnObjToDec$ will have one connected component while the $\knnDecToObj$ has multiple components. If you have an easy problem you would expect points in the decision space and their image in the objective space to be close to each other. Thus, the distances of the points after edge transfer will be small. For a different problem, the points in the decision space might be close to each other but not in the objective space which will be reflected in the graphs.
Another example is displayed in Figure \ref{fig:mo-ela-visualisation}. The edges of the MST $\stDec$ are applied to the objective space. The created $\stDecToObj$ is not an MST any more. Additionally, you can observe that e.g., the angle spanned by the points differ drastically. 
These differences as well as the general structure of the decision and objective space is of interest when characterising a problem.

First, we describe the features applied to both graphs. The minimum, maximum, and average distance between nodes (\texttt{weights\_(min|max|avg)}) are used as features. For every node in a graph the closeness centrality is computed. Closeness centrality quantifies the average shortest path length from a node to all other nodes in a network. From that, the lowest, highest and average closeness centrality (\texttt{closeness\_centrality\_(min|max|avg)}) is used as a feature. Additionally, if a node is connected to two other nodes the angle between these two nodes is calculated, see angles $\alpha_x$ and $\alpha_y$ in Figure \ref{fig:mo-ela-visualisation}. Here, the minimum, maximum, and average of all angles are used as features (\texttt{angle\_(min|max|avg)}). These angle features are inspired by the angle features used in single-objective ELA (SO-ELA)~(\cite{kerschke2014cell}), which proved to perform well.
The following features are only calculated on the $k$NN based graphs: The total number of connected components (\texttt{num\_components}), the min, max, and average nodes per connected component (\texttt{(min|max|avg)\_nodes\_per\_component}) and the longest path per component is computed and the min, max, and average distance of all paths (\texttt{longest\_path\_(min|max|avg)}) are used as features. 
For the MST the distance of the longest path (\texttt{longest\_path}) is used directly as a feature. An addition to all these features the ratio of the features given by $\knnDec$ and $\knnObj$ as well as $\stObj$ and $\stDec$ are calculated.
        
In the case of $|L_1| = 1$, i.e., there is only one non-dominated point, only a graph with one node and no edges can be created. Thus, features relying on distances or angles can not be computed and are set to~$0$. The features describing the number of connected components are set to~1.
In total we created $50$ features based on $\stGraph$ and 80 based on $\knnGraph$.

\subsection{Gradient-Based Features}

The last feature set is based on the gradient of the non-dominated points \(L_1\) in the sample. The MST \(\stObj\) is calculated first.
Next, for each edge \(\{i,j\} \in E(\stObj)\), each objective \(k=1,\ldots,\objSpaceDim\), and each decision variable \(p=1,\ldots,\decSpaceDim\), we compute absolute per-variable slopes
\[
r_{k,p}(i,j) \;=\; \frac{\lvert f_k(x_i) - f_k(x_j) \rvert}{\lvert x_{i,p} - x_{j,p} \rvert}.
\]
Concatenating these slopes over all edges yields, for each objective \(k\), a vector \(g_k \in \R^{\decSpaceDim \cdot \lvert E(\stObj)\rvert}\). These are then combined into a single multi-objective gradient by averaging across objectives,
\[
\bar g \;=\; \frac{1}{\objSpaceDim}\sum_{k=1}^{\objSpaceDim} g_k \;\in\; \R^{\decSpaceDim \cdot \lvert E(\stObj)\rvert}.
\]
As features we use the elementwise minimum, maximum, standard deviation, and average of \(\bar g\) (\texttt{mo\_gradient(min|max|std|avg)}), totaling $4$ features. If fewer than two points are available (\(\decSampleSize \le 1\)), all four features are set to \(0\).

\begin{table}
    \caption{Overview of the proposed features. Features already proposed in~\cite{LieEtAl2021} are marked with an asterisk (*).}
    \label{tab:all_features}
    \centering
    \renewcommand{\tabcolsep}{1pt}
    \renewcommand{\arraystretch}{0.85}
        \begin{tabularx}{\textwidth}{clXr}
        % \toprule
        & \textbf{Feature} & \textbf{Description} & \textbf{}\\
        \midrule
        \multirow{6}{*}{\rotatebox[origin=c]{90}{\textbf{NDS}}} 
        & $\text{no\_non\_dom\_points}^*$ & number of non-dominated points & \\
        & $\text{max\_rank}^*$ & number of domination layers & \\
        & $\text{avg\_points\_per\_layer}^*$ & average number of points per domination layer & \\
        & hv\_dom\_layer\_$i$ & HV in objective space of domination layer $i$ & \\
        & sp\_dom\_layer\_$i$ & SP in decision space of domination layer $i$ & \\
        & r$p$ & the $R^2$ indicator for polynomial degree $p$ & \\
        
        \midrule
        \multirow{8}{*}{\rotatebox[origin=c]{90}{\textbf{Descriptive}}} 
        & min\_$m$ & minimum of objective $m$ & \\
        & max\_$m$ & maximum of objective $m$ & \\
        & avg\_$m$ & average of objective $m$ & \\
        & std\_$m$ & standard deviation of objective $m$ & \\
        & obj\_std\_diff & the absolute difference of the standard deviations of the objectives (only for $m=2$) & \\
        & $\text{spearman\_corr\_obj}^*$ & The (Spearman) correlation of the objectives (only for $m=2$) & \\
        
        \midrule
        \multirow{6}{*}{\rotatebox[origin=c]{90}{\textbf{PCA}}} 
        & (min$\vert$max$\vert$avg$\vert$std)\_pc\_$X$ & the minimum, maximum, average and standard deviation of the variance explained by all PCs for $X$ & \\
        & (min$\vert$max$\vert$avg$\vert$std)\_pc\_$y$ & the minimum, maximum, average and standard deviation of the variance explained by all PCs for $y$ & \\
        & (min$\vert$max$\vert$avg$\vert$std)\_pc\_$X\_y$ & the minimum, maximum, average and standard deviation of the variance explained by all PCs for $X\_y$ & \\
        
        \midrule
        \multirow{24}{*}{\rotatebox[origin=c]{90}{\textbf{Graphs}}} 
        & weights\_min & minimum distance of all connections & \\
        & weights\_max & maximum distance of all connections & \\
        & weights\_avg & average distance of all connections & \\
        & closeness\_centrality\_min & minimum closeness centrality of all nodes & \\
        & closeness\_centrality\_max & maximum closeness centrality of all nodes & \\
        & closeness\_centrality\_avg & average closeness centrality of all nodes & \\
        & angle\_centrality\_min & minimum angle between two nodes & \\
        & angle\_centrality\_max & maximum angle between two nodes & \\
        & angle\_centrality\_avg & average angle between two nodes & \\

        & \makecell[l]{\raggedright \textit{NN specific}} & & \\
        & num\_components & total number of connected components &  \\
        & nodes\_per\_components\_min & minimum number of nodes in a connected component & \\
        & nodes\_per\_components\_max & maximum number of nodes in a connected component & \\
        & nodes\_per\_components\_avg & average number of nodes in a connected component & \\
        & longest\_path\_components\_min & the shortest path of the longest path of all connected components & \\
        & longest\_path\_components\_max & the longest path of the longest path of all connected components & \\
        & longest\_path\_components\_avg & the average path of the longest path of all connected components & \\

        & \makecell[l]{\raggedright \textit{MST specific}} & & \\
        & longest\_path & the longest path & \\
        
        \midrule
        \multirow{4}{*}{\rotatebox[origin=c]{90}{\textbf{Gradient}}} 
        & mo\_gradient\_min & minimum multi-objective gradient & \\
        & mo\_gradient\_max & maximum multi-objective gradient & \\
        & mo\_gradient\_avg & average multi-objective gradient & \\
        & mo\_gradient\_std & standard deviation multi-objective gradient & \\
        
        % \bottomrule
    \end{tabularx}
\end{table}

\section{Experiments}
\label{sec:experiments}

This section outlines the experiments conducted to evaluate our proposed features. We do this by using them for an automated algorithm selection task, followed by feature selection to identify the most relevant MO-ELA features. The selected set of features performs very effectively, enabling clear distinction between different types of optimization problems. Furthermore, we show that these features are complementary and consistent, highlighting their robustness and value for characterizing problem landscapes.

\subsection{Setup}
\label{subsec:setup}

\paragraph{Benchmark Set} Our benchmark set is a compilation of BiObjBBOB~(\cite{BrockhoffEtAl22BiObjBBOB}), ZDT~(\cite{zitzlerPerformanceAssessmentMultiobjective2003}), DTLZ~(\cite{debScalableTestProblems2005}), and MPM2~(\cite{schapermeierPeakABooGeneratingMultiobjective2023}), four well-known benchmark sets used in multi-objective optimisation. We consider functions with $\objSpaceDim \in \{2, 3\}$ and $\decSpaceDim \in \{2, 5, 10, 20\}$ 

For the bi-objective functions we took all $92$ functions with its first instance from BiObjBBOB, from ZDT we took functions $1, 2, 3, 4, \text{ and } 6$, from DTLZ we took all $7$ functions, and from MPM2 we combined two single-objective functions with $peaks \in \{2^k \mid k = 0, 1, \ldots, 7\}$, $topology \in \{\text{random}, \text{funnel}\}$ and took all distinct combinations for each objective without considering the order which results in $72$ functions. Considering all functions over all $\decSpaceDim$ we get in total $704$ bi-objective functions.

For the tri-objective functions, we chose all $7$ functions from ZDT, but only considered $\decSpaceDim \geq \objSpaceDim$. From MPM2 we combined three single-objective functions with $peaks \in \{2^k \mid k = 0, 1, \ldots, 5\}$, $topology \in \{\text{random}, \text{funnel}\}$ and used all distinct combinations for each objective again without considering the order which results in $110$ functions. Considering all $\decSpaceDim$ we created $461$ tri-objective functions.
In total we thus have $1\,165$ functions.
    
\paragraph{MO-Algorithms} The algorithms used to solve the given problems are the basis for the algorithm selection. We chose three commonly-used multi-objective solvers namely \NSGAII~(\cite{debFastElitistMultiobjective2002}), \SMSEMOA~(\cite{BeumeEtAl07}), and MOEA/D~(\cite{MOEAMultiobjectiveEvolutionary2007}). For \NSGAII and \SMSEMOA we used the \texttt{R} package \texttt{ecr}~(\cite{Bossek2017ecr2}) and for MOEA/D we used the package \texttt{moeadr}~(\cite{CamBatAra2020moeadr}). For each solver the given default configuration of the respective package was chosen.

\paragraph{Performance Data}  For each of the three solvers we performed $20$ runs on every instance with $100\,000$ evaluations. Out of the resulting 60 runs per instance, we took the set of non-dominated points, extracted the maximum value per objective, and multiplied it by $1.1$ to construct the reference point~(\cite{ishibuchi2018specify}). This reference point is used to calculate the HV of a solved instance. In addition, we calculated a reference HV based on all extracted non-dominated points and the reference point. \\
Table \ref{tab:function_evals} shows the number of evaluations used for assessing the final performance of a solver run. The number of evaluations is dependent on $\decSpaceDim$ and $\objSpaceDim$ and was determined by observing the convergence of the previously mentioned solver runs.
The final performance per algorithm on each of the instances is the arithmetic mean of the normalised HV (HVN) over $20$ runs on the final populations of the solvers. The HVN is calculated by dividing the achieved HV of the final population by the reference HV, i.e., the HV obtain with the reference set. This ensures that the maximum HV equals $1$ and the performance of the solvers can be compared across instances.

\begin{table}
    \centering
    \caption{Number of evaluations used per combination of $d$ and $m$ for a solver run.}
    \renewcommand{\tabcolsep}{13pt}
    \label{tab:function_evals}
    \begin{tabular}{r|rrrr}
        $\objSpaceDim$ / $\decSpaceDim$ & $2$ & $5$ & $10$ & $20$ \\
        \midrule
        $2$ & $3,000$ & $8,000$ & $15,000$ & $20,000$ \\
        $3$ & $7,000$ & $25,000$ & $50,000$ & $100,000$
    \end{tabular}
\end{table}

\paragraph{Feature Calculation} For the feature calculation different samples of size $s \in \{100, 200, 500, 1\,000\}$ with $20$ different seeds using Latin hypercube sampling (LHS) for every instance of our benchmark set were created. This resulted in $1\,165$ instances $\times$ $4$ different sample sizes $\times$ $20$ seeds = $93\,200$ feature calculations. 
For all problems, all existing features of \cite{LieEtAl2021} and our novel features were calculated. That resulted in $53 + 226 = 279$ features for the bi-objective problems and $59 + 233 = 292$ features for the tri-objective problems.

\paragraph{Algorithm Selectors} We chose three machine learning algorithms for an algorithm selector, a Support Vector Machine (SVM), eXtreme Gradient Boosting (XGBoost), and a Random Forest (RF). For the SVM and RF we used the package \texttt{sklearn}~(\cite{scikit-learn}) and for XGBoost we chose the \texttt{XGBoost}~(\cite{ChenEtAl16XGBoost}) package. The SVM was used with default parameters, for XGBoost and RF, we chose $500$ and $200$ trees respectively while keeping the remaining parameters default. The reduced estimators for the RF are due to runtime restrictions caused by the extensive feature selection described in the next paragraph.

\paragraph{Feature Selection}
Forward-backward feature selection was used to perform feature selection, using the package \texttt{mlxtend}~(\cite{raschkas_2018_mlxtend}). The forward-backward feature selection is an exhaustive technique for calculating high performing feature combinations from the given set. It starts with a forward selection where it uses an empty set of features and successively trains a model with one feature and selects the feature that improves the model performance the most. After that, a second feature is added. As soon as there are more than two features a backward selection after every forward selection is performed. All features chosen so far are successively removed from the current features set and the models' performance is tested. If the performance does not decrease the feature is not considered as a contributing feature any more. 
As a performance metric, we chose the F1-score due to unbalanced classes caused by some solvers being better on more of the chosen problems.
In our opinion this exhaustive feature selection is important to get insights into the best-performing complementary combination of features that yield the best performance
out of the multitude of features proposed in this paper. Although it has to be mentioned that feature selection can be multi-modal depending on the given setting. Thus, there can be different feature combinations with an equal or better performance than found by the forward-backward feature selection.

\paragraph{Automated Algorithm Selection}

\begin{figure}
     \centering
     \includegraphics[width=0.8\textwidth]{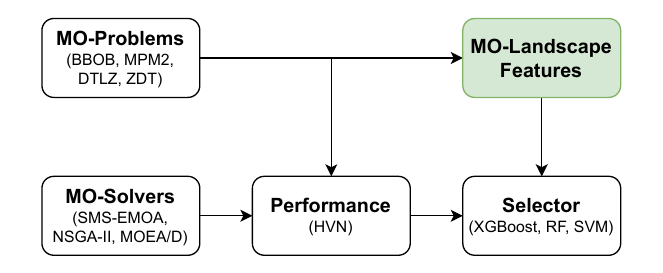}
    \caption{Overview of the algorithm selection training framework for multi-objective problems. The green indicated box highlights our contribution. }
    \label{fig:selection_overview}
\end{figure}

The data set used for the AAS and feature selection consists of the feature vectors calculated based on all function, dimension, and sample size combinations. Since we have a data set of feature vectors of problems of different sample sizes and dimensions we used the sample size and dimension as additional features. The solver with the highest mean HVN over $20$ runs is used as a label. 
The AAS is performed independently for the bi- and tri-objective problems
resulting in two AAS scenarios. The resulting data sets are each split into an $80/20$ random train/test split. We decided against cross-validation as we have a large enough data set that supports results with a standard train/test split and cross-validation would add additional computational cost to our already large experiment.
With the given data sets we trained all three ML algorithms in combination with the forward-backward feature selection. On each step of the feature selection, a $3$-fold cross-validation was used.
This resulted in three selectors each for the bi-objective and tri-objective functions. To determine which of the three selectors performs best, we need to calculate how much the VBS / SBS gap was closed. The virtual best solver (VBS) describes the solver which is always the best for a specific problem, i.e. the label that is used for training a selector. The single best solver (SBS) describes the solver which solves the most problems as the best solver. The VBS / SBS gap describes how much performance improvement can be gained by always selecting the best solver (VBS) for a specific instance compared to the solver best across all problems (SBS).
Closing this gap can be calculated with the relative improvement (RI) = (selected\_HVN - SBS\_HVN) / (VBS\_HVN - SBS\_HVN) where a RI of $1$ closes the gap completely, i.e. always choosing the VBS and with a RI of $0$ the SBS was always chosen. When the RI is negative the selector chooses algorithms that are worse than the SBS. Note here that negative values can be smaller than $-1$. This happens when the performance difference of a bad solver to the SBS is larger than the performance difference of the SBS to the VBS.
The selector with the highest RI on the test set was used as a final selector resulting in one selector for bi-objective functions and one for tri-objective functions. 
Figure \ref{fig:selection_overview} shows an overview of the algorithm selection pipeline with the used setting. The difference and contribution of this paper are the MO-Landscape features highlighted in green.

We are using this AAS study as a proof-of-concept to showcase the expressiveness and of the newly developed features and their complementariness to already existing multi-objective features. Thus, the used selectors and solvers are limited. Though, the machine learning algorithms for the selector are standard algorithms that prove to work well in SO-AAC. Further, are the three solvers well established algorithms which are standard evolutionary solvers that prove to work well in various optimisation settings.

\subsection{Results}
\label{results}

\begin{figure}
    \begin{subfigure}[b]{\textwidth}
         \centering
         \includegraphics[width=\textwidth]{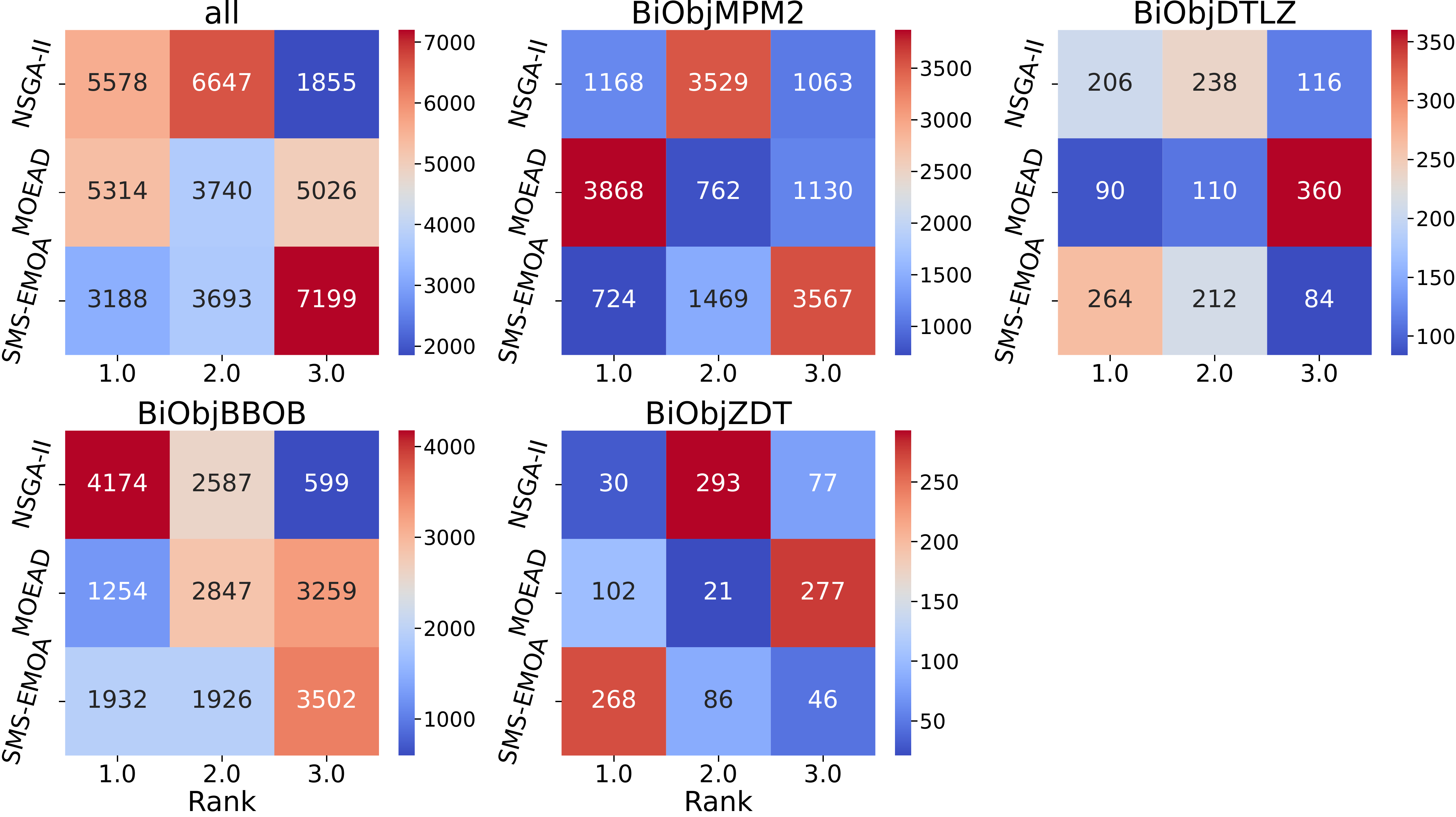}
         \caption{Contingency tables for 2 objectives.}
         \label{fig:contingency_tables_2_obj}
     \end{subfigure}
    \begin{subfigure}[b]{\textwidth}
         \centering
         \includegraphics[width=\textwidth]{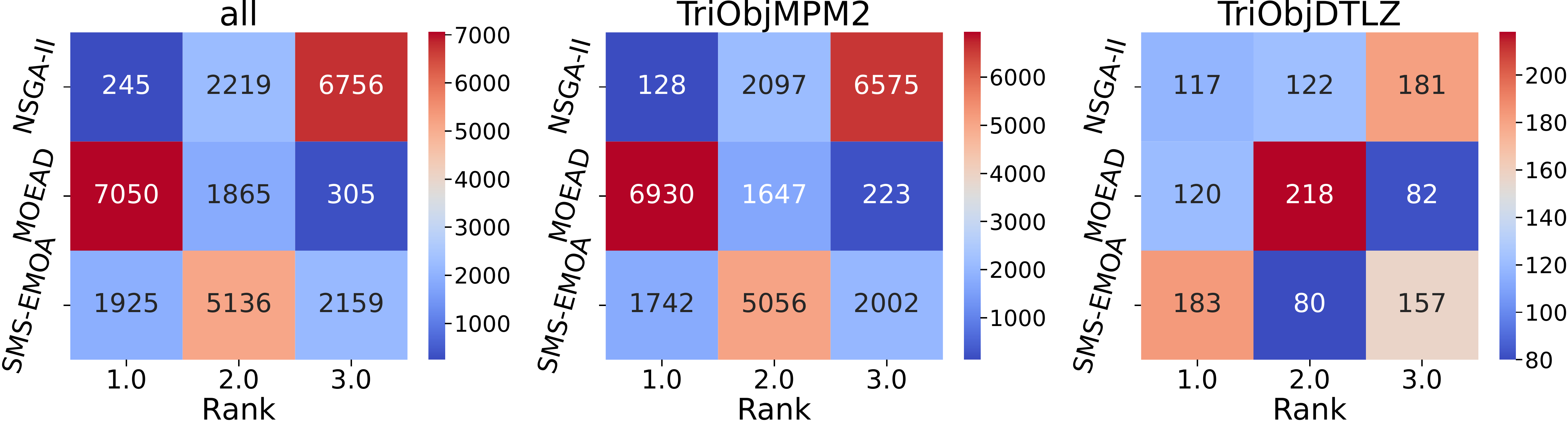}
         \caption{Contingency tables for 3 objectives.}
         \label{fig:contingency_tables_3_obj}
     \end{subfigure}
    \caption{Contingency table for the different combinations of benchmark, dimension and number of objectives.}
    \label{fig:contingency_tables}
\end{figure}

\paragraph{Automated Algorithm Selection}
First, we investigate the results of the AAS. Figure~\ref{fig:contingency_tables} shows a contingency table of the ranks achieved by the solvers on the different and combined benchmark sets displaying the need for AAS. The ranks are based on all runs with the same starting seed. This results in 704 bi-objective problems $\times$ $20$ runs = $14,080$ rankings and in 461 tri-objective problems $\times$ $20$ runs = $9,220$ rankings. For the bi-objective problems we see that \NSGAII is the best performing solver on the combined benchmark set closely followed by \MOEAD with both solvers winning on about $5,500$ ($39\%$) runs \SMSEMOA ranks first on about $3,200$ ($22\%$) runs. We further see that \NSGAII wins most of the runs for BiObjBBOB and \MOEAD on BiObjMPM2 which explains their almost tied first ranks on the combined benchmark set since BiObjBBOB and BiObjMPM2 are the largest benchmark sets. \SMSEMOA gains its most first ranks on BiObjBBOB and wins BiObjDTLZ and BiObjZDT, although these are small benchmark sets compared to the others. For the tri-objective problems \MOEAD wins the majority of the runs with about $7,000$ ($76\%$) first ranks. \SMSEMOA ranks second with about $1,900$ ($21\%$) first ranks and \NSGAII ranks third with $245$ ($3\%$) first ranks. When looking at the ranks of TriObjDTLZ we see that the ranks follow closely an equal distribution while \SMSEMOA still has the most first ranks. The more uneven distribution of first ranks is caused by TriObjMPM2 where \MOEAD ranks first on $78\%$ of the instances. 

\begin{figure}
    \centering
    \includegraphics[width=0.45\linewidth]{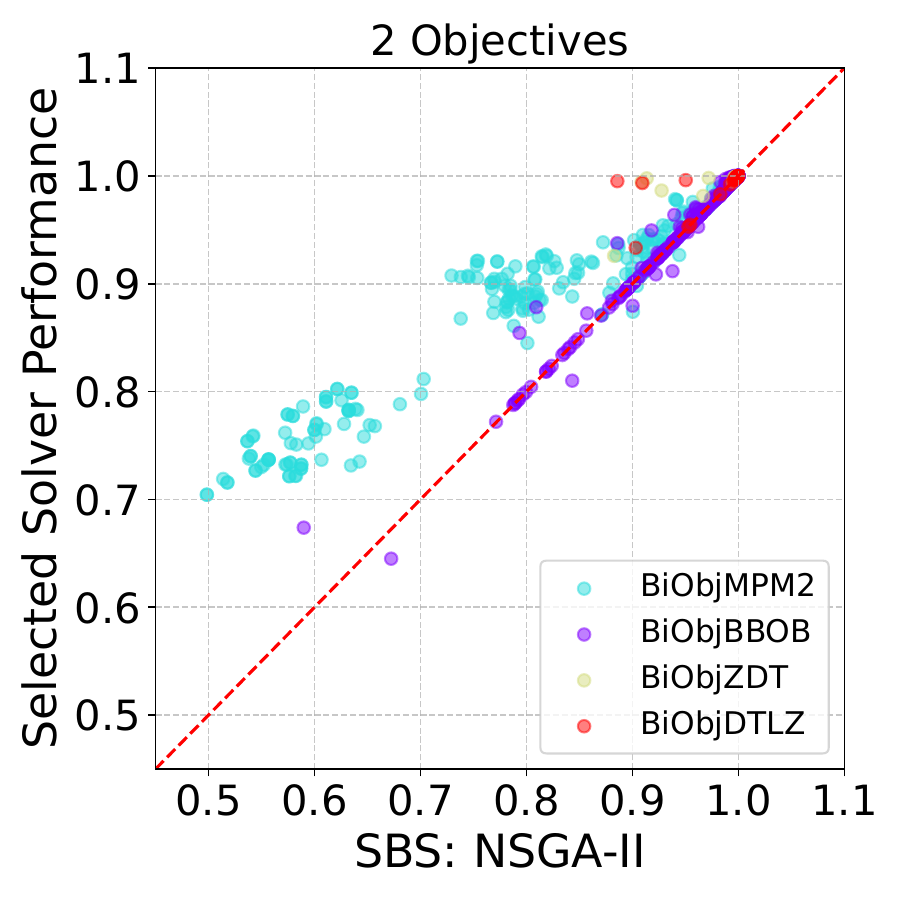}
    ~
    \includegraphics[width=0.45\linewidth]{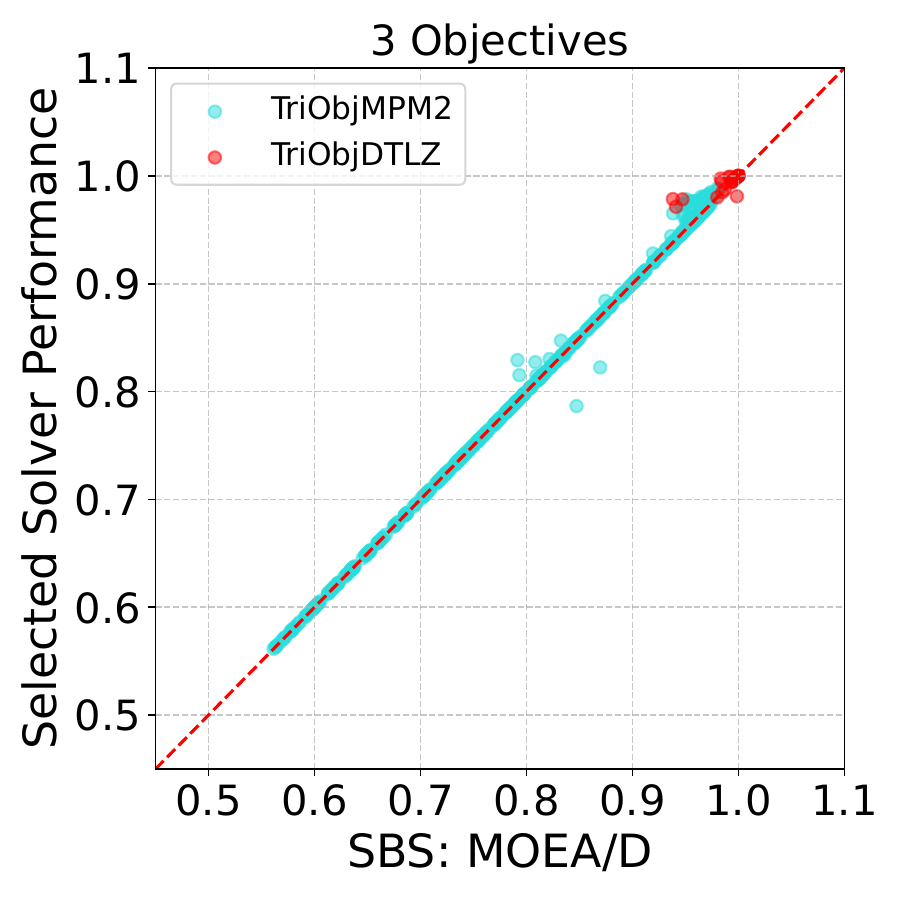}
    \caption{Comparison of the predictions of the best selector against the SBS on the test set containing problems of all dimensions and sample sizes.}
    \label{fig:sbs_vs_best_selector}
\end{figure}

These rankings show that there is potential for improvement using AAS. To show this we compare three selection scenarios. The first is an XGBoost combined with the feature selection, as the XGBoost achieved the highest RI for both the bi- and tri-objective functions. The second is an XGBoost without feature selection as a baseline and as a  third scenario we chose a random selector. 
You can see that the random selector has an F1 score of $0.31$ and $0.24$ for the bi- and tri-objective functions respectively, when considering all benchmark sets. Combined with a negative RI of $-0.12$ and $-24.74$ for the bi- and tri-objective functions you can clearly see that the random selector not only performs the worst but also performs worse than always choosing the SBS.
Compared to that increases the baseline XGBoost the performance by a large amount. It achieves an F1 of $0.928$ for the bi-objective and $0.959$ for the tri-objective functions. 
The RI achieves values of $0.961$ and $0.262$   respectively for the bi- and tri-objective functions. The XGBoost combined with a feature selection increases the values further. The F1 increases to $0.941$ and $0.967$ while the RI increases to $0.963$ and $0.272$ for the for the bi- and tri-objective functions respectively.
A more detailed analysis based on the different benchmark sets can be found in Tab. \ref{tab:improvements}. \\
Figure~\ref{fig:sbs_vs_best_selector} shows the actual HVN of the SBS compared to the selected solver. Points above the diagonal line indicate a better performance over the SBS. You can see especially an improvement for the bi-objective MPM2 problems. In the figure for the tri-objective problems we see a small absolute HVN improvement of the selected solver over the SBS. This is because \MOEAD is on a majority of the problems the best solver and selecting \MOEAD if \MOEAD is the VBS does not improve the RI or shows a point above the line.

\begin{table}
    \caption{Performance comparison of F1 and relative improvement (RI) across different benchmarks. Columns indicated with $r$ show performance of a random selector, columns with $b$ show the baseline performance without feature selection, and columns indicated with $s$ show the performance with feature selection.}
    \centering
    \footnotesize
    \label{tab:improvements}
    \begin{tabular}{lc|ccc|ccc|ll}
    $\objSpaceDim$ & \textbf{selector} & \textbf{$\text{F1}_r$} & \textbf{$\text{F1}_b$} & \textbf{$\text{F1}_s$} & \textbf{$\text{RI}_r$} & \textbf{$\text{RI}_b$}  & \textbf{$\text{RI}_s$} & \textbf{SBS} & \textbf{benchmark} \\
    \midrule
     &          & $0.303$ & $0.957$ & $\mathbf{0.965}$ & $-1.391$ & $0.933$ & $\mathbf{0.940}$ & \NSGAII & DTLZ \\
     &          & $0.278$  & $\mathbf{0.948}$ & $0.941$ & $0.202$ & $\mathbf{0.985}$  & $0.985$ & \MOEAD & MPM2 \\
     $2$ & XGBoost & $0.197$ &  $0.977$  & $\mathbf{0.984}$ & $-0.054$ & $\mathbf{0.982}$  & $0.980$ & \SMSEMOA & ZDT \\
     &          & $0.293$ & $0.875$  & $\mathbf{0.904}$ & $-8.831$ & $0.383$  & $\mathbf{0.446}$ & \NSGAII & BBOB \\
     &          & $0.313$ & $0.928$  & $\mathbf{0.941}$ & $-0.126$ & $0.961$  & $\mathbf{0.963}$ & \NSGAII & all \\
     \midrule
     &          & $0.230$ & $0.938$  & $\mathbf{0.953}$ & $-30.214$ & $-0.127$ & $\mathbf{-0.110}$ & \MOEAD & MPM2 \\
     $3$ & XGBoost & $0.321$ & $0.993$  & $\mathbf{0.993}$ & -$1.647$ & $0.546$  & $\mathbf{0.550}$ & \MOEAD & DTLZ \\
     &          & $0.240$ & $0.959$  & $\mathbf{0.967}$ & -$24.749$ & $0.262$  & $\mathbf{0.272}$ & \MOEAD & all \\
    \end{tabular}
\end{table}

\paragraph{Feature Selection}
The feature selection integrated into the AAS shows that the number of features for the given setting can be reduced drastically. Out of the total $226$ features $32$ were ultimately chosen for the bi-objective functions. Of these $32$ features $14$ were selected from the newly proposed features and 18 from the features proposed by \cite{LieEtAl2021}. The most common feature group of the new features is NDS with $7$ selected features, the feature groups MST and NN have $3$ and $2$ features respectively and from PCA and Gradient each $1$ feature was selected. The most common feature group of \cite{LieEtAl2021} is `global' with $7$ features, followed by `evolvability' with 6 features. `Multimodality' and `ruggedness' have $3$ and $2$ features respectively. 
With NDS, which broadly captures the `global' structure of a problem, and global being the most selected feature groups it shows that the global structure is the most important aspect to choose a correct solver for the performed AAS.
Neither the dimension nor the sample size was chosen as a feature for the bi-objective functions. This indicates that the constructed features themselves already contain this information (Tab.~\ref{tab:feature_groups}). 

Out of all $233$ features $36$ were optimally chosen for the tri-objective functions. Out of these were $17$ of the newly proposed features and $18$ of the features proposed by \cite{LieEtAl2021}. The most common feature group of the newly created features was NN with $11$ features. MST was chosen with $3$ features and the remaining feature groups NDS, PCA and Gradient each contributed $1$ feature. The feature groups `global', `evolvability' and `ruggedness' from \cite{LieEtAl2021} were each selected $5$ times and `multimodality' $3$ times. The chosen feature groups show that the local structure of tri-objective problems is the most important aspect to look for in the performed AAS with NN mostly capturing the local structure of problems. That said it has to be mentioned that the \cite{LieEtAl2021}'s feature groups `global', `evolvability' and `ruggedness' also contribute a large part of the features. This indicates that different properties of tri-objective functions need to be considered as well.
In addition, the feature selection determined the dimension as an important feature. 
Worth noting here is that none of the features of the Descriptive feature group was chosen for the bi- or tri-objective problems (Tab.~\ref{tab:feature_groups}). However, these features could have possibly been chosen in a different combination as feature selection is multi-modal where multiple different feature combinations can potentially lead to the same performance.

\begin{table*}
    \centering
    \footnotesize
    \caption{Feature group distribution for bi-objective and tri-objective problems.}
    \label{tab:feature_groups}
    \renewcommand{\tabcolsep}{2.5pt}
    \renewcommand{\arraystretch}{.85}
    \begin{tabular}{r|cccccc|cccc|cc|c}
      $m$ & NDS & MST & NN & PCA & Grad. & Descr. & multimod. & global & evolv. & rugged. & dim & sample size & sum \\
    \midrule
    $2$ & $7$ & $3$ & $2$ & $1$ & $1$ & $0$ & $3$ & $7$ & $6$ & $2$ & $0$ & $0$ & $32$ \\
    $3$ & $1$ & $3$ & $11$ & $1$ & $1$ & $0$ &  $3$ & $5$ & $5$ & $5$ & $1$ & $0$ & $36$ \\
    \end{tabular}
\end{table*}

\begin{figure}
    \centering
    \includegraphics[width=0.49\linewidth]{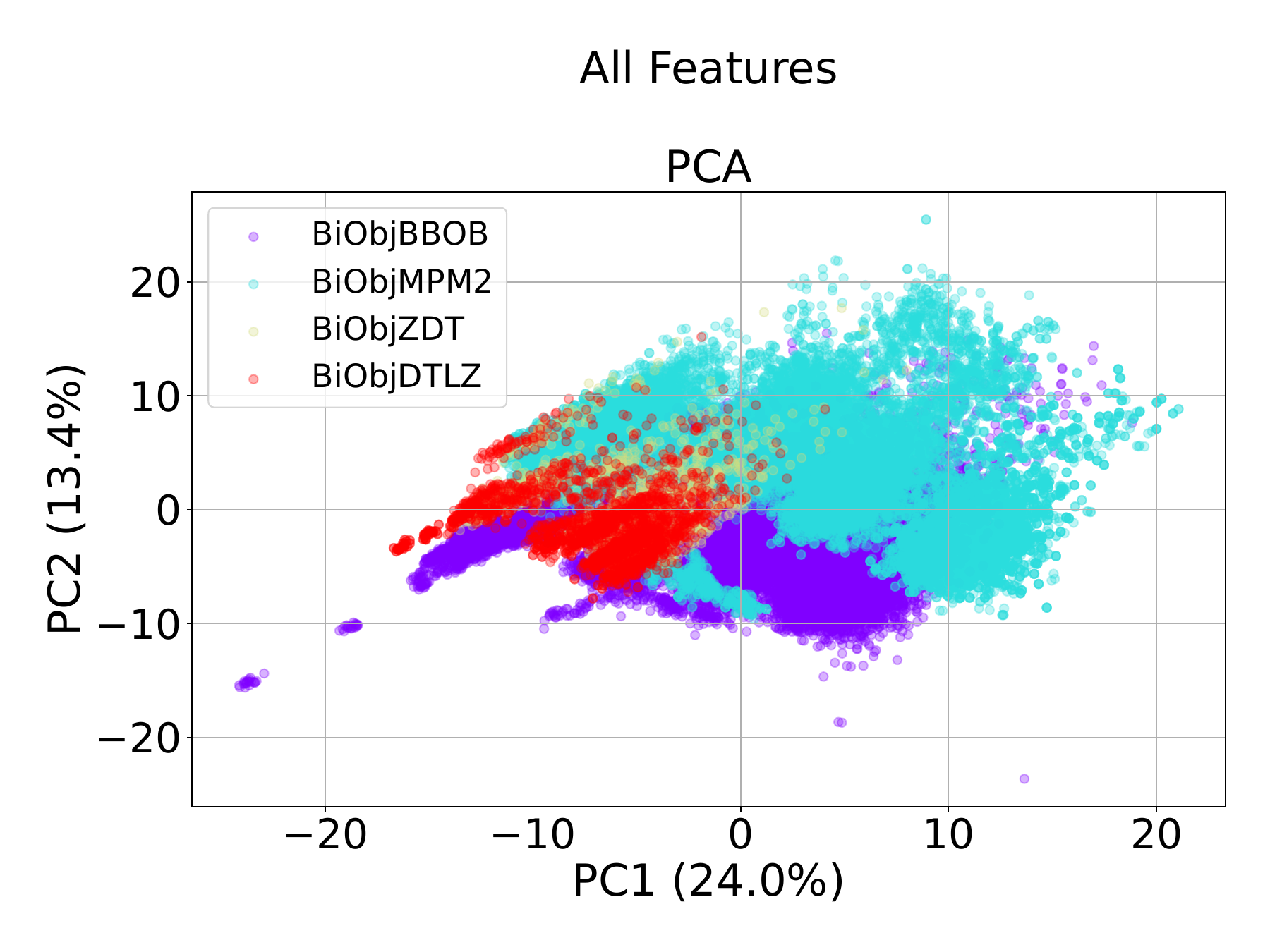}
    \includegraphics[width=0.49\linewidth]{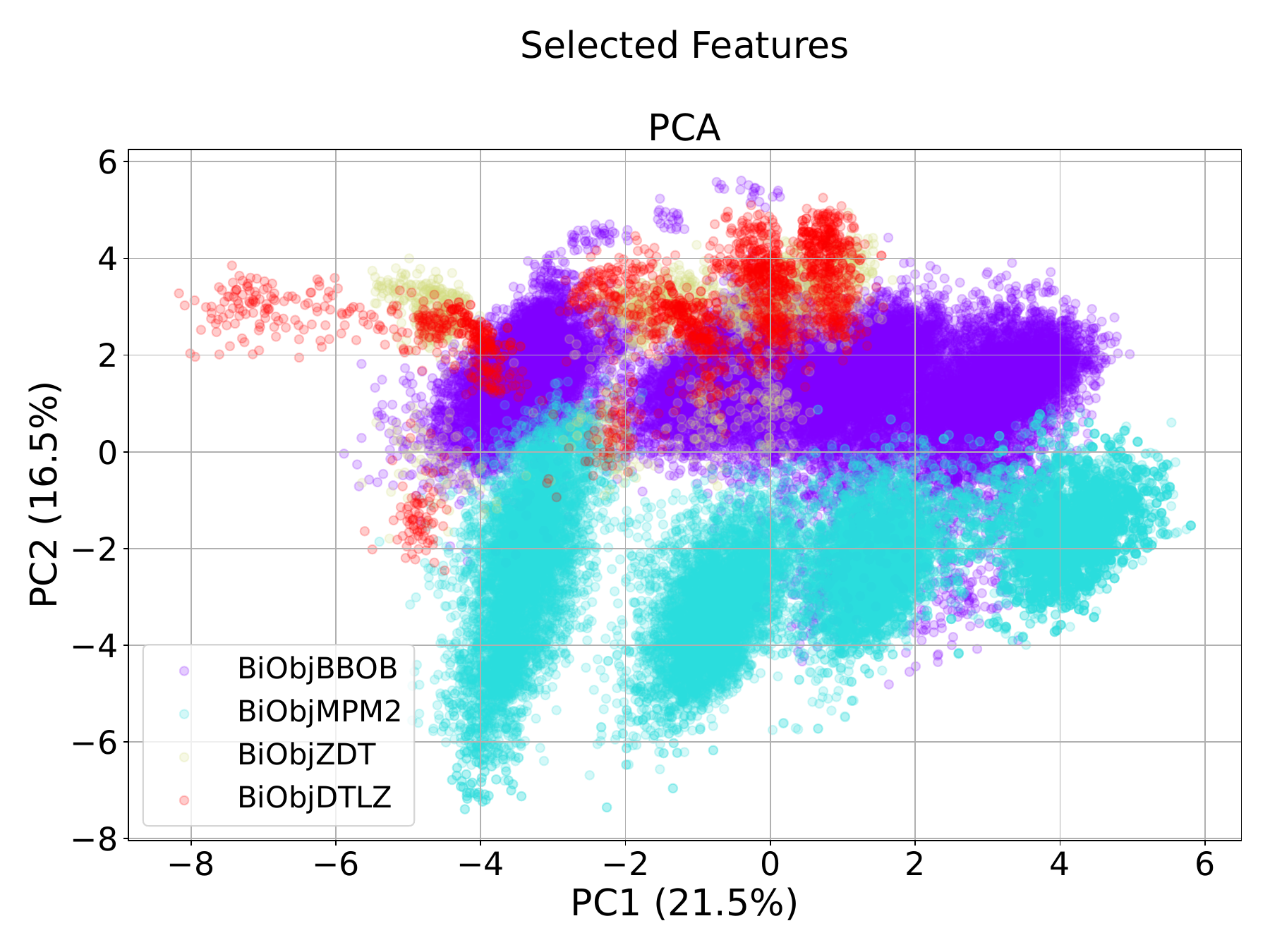}
    \includegraphics[width=0.49\linewidth]{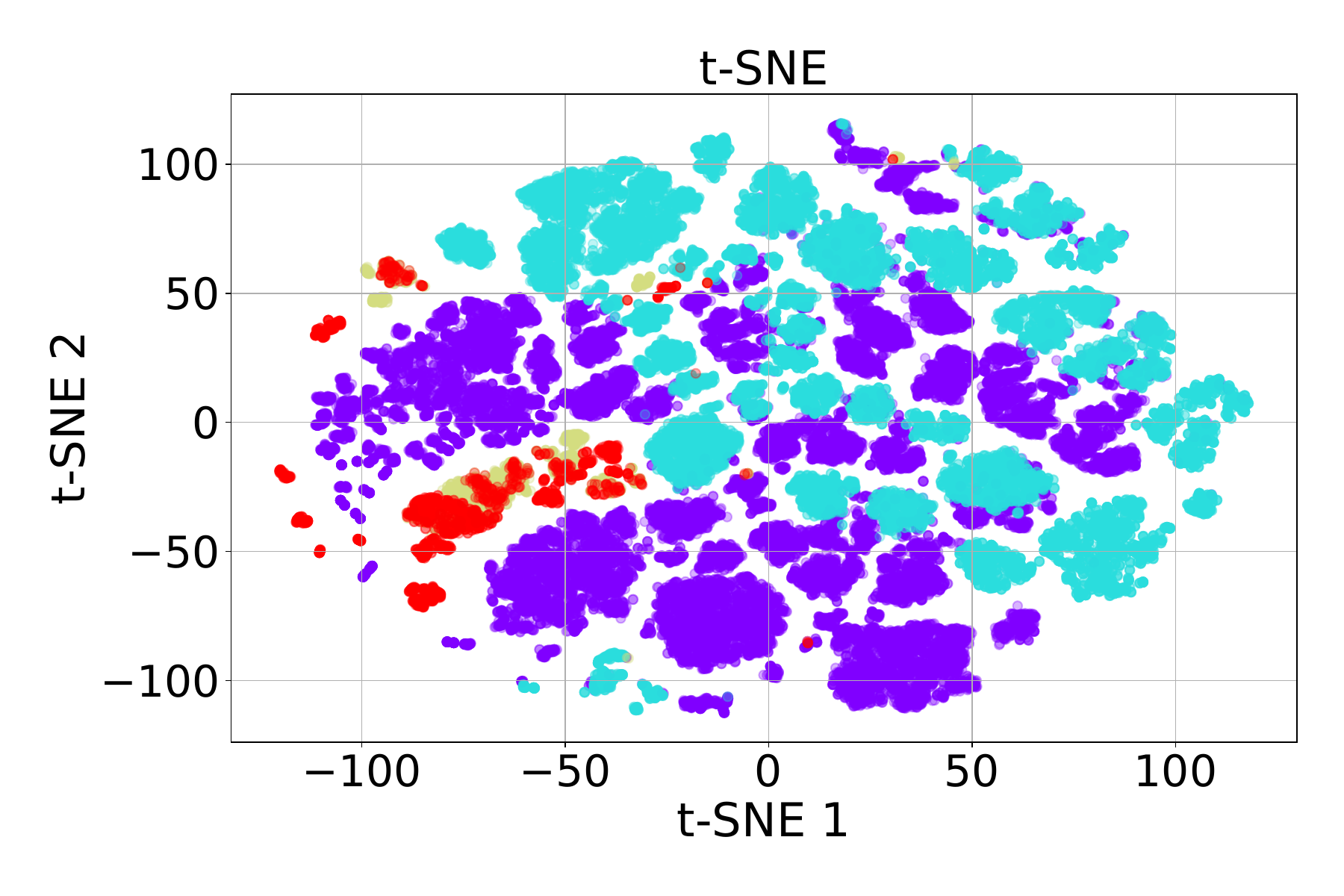}
    \includegraphics[width=0.49\linewidth]{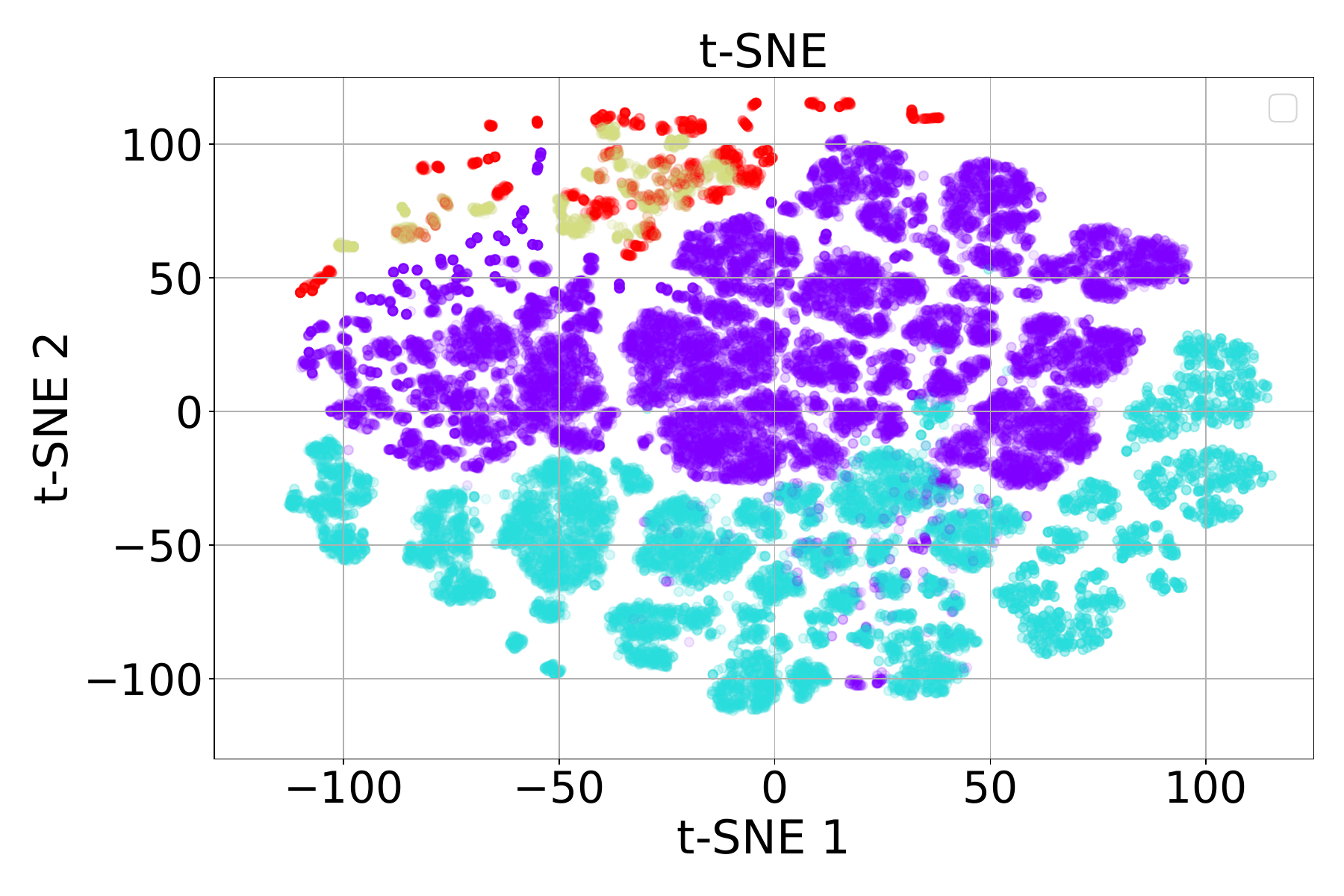}
    \caption{Dimensionality reduction using PCA and t-SNE for the bi-objective problems. The left plot uses all features. The right plot uses the selected features.}
    \label{fig:pca_by_benchmark_2obj}
\end{figure}

\paragraph{Visualisation}
We now focus more on the features themselves. For that, we took all feature vectors for the bi- and tri-objective functions separately and applied a principal component analysis (PCA) and t-distributed stochastic neighbour embedding (t-SNE)~(\cite{tsne}) with a perplexity value of $30$. PCA and t-SNE are dimensionality reduction techniques which allow high dimensional data to be reduced to two dimensions to be visualised in e.g. scatter plots. Such scatter plots can be seen in Figure~\ref{fig:pca_by_benchmark_2obj} for the bi-objective problems. The left column of Figure~\ref{fig:pca_by_benchmark_2obj} shows the dimensionality reduction based on all features. For both PCA and t-SNE you can not see a clear structure in the data. The different benchmark sets overlap and do not form any clear clusters. To further investigate this we repeated the dimensionality reduction with only the $34$ chosen features of the feature selection for the bi-objective problems (right column). Reducing the features to the most important ones reduced the noise in the data in a way that a separation between the benchmark sets becomes visible. With PCA BiObjBBOB and BiObjMPM2 form two large clusters with some overlap between the problems. The problems overlapping from BiObjBBOB are multi-modal. This is not surprising since BiObjMPM2 is highly multi-modal. BiObjDTLZ and BiObjZDT have a clear overlap with each other since both are constructed to have similar properties. These properties seem to be shared with some functions from BiObjBBOB since there is also a clear overlap. Another separation that can be seen are four clusters within the benchmark sets, especially BiObjMPM2, BiObjDTLZ and BiObjZDT which represent the different decision space dimensions.

\begin{figure}
    \centering
    \includegraphics[width=0.49\linewidth]{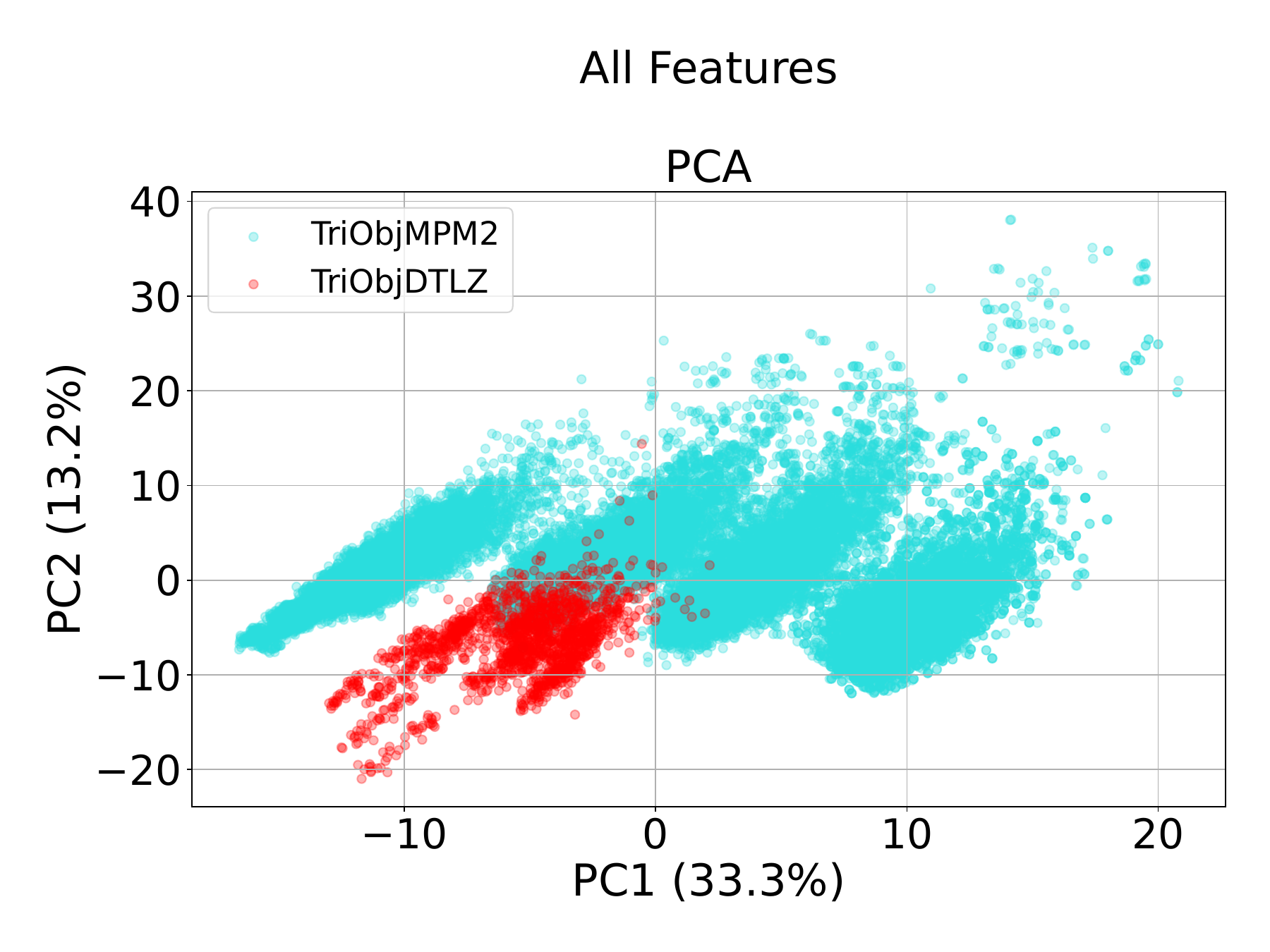}
    \includegraphics[width=0.49\linewidth]{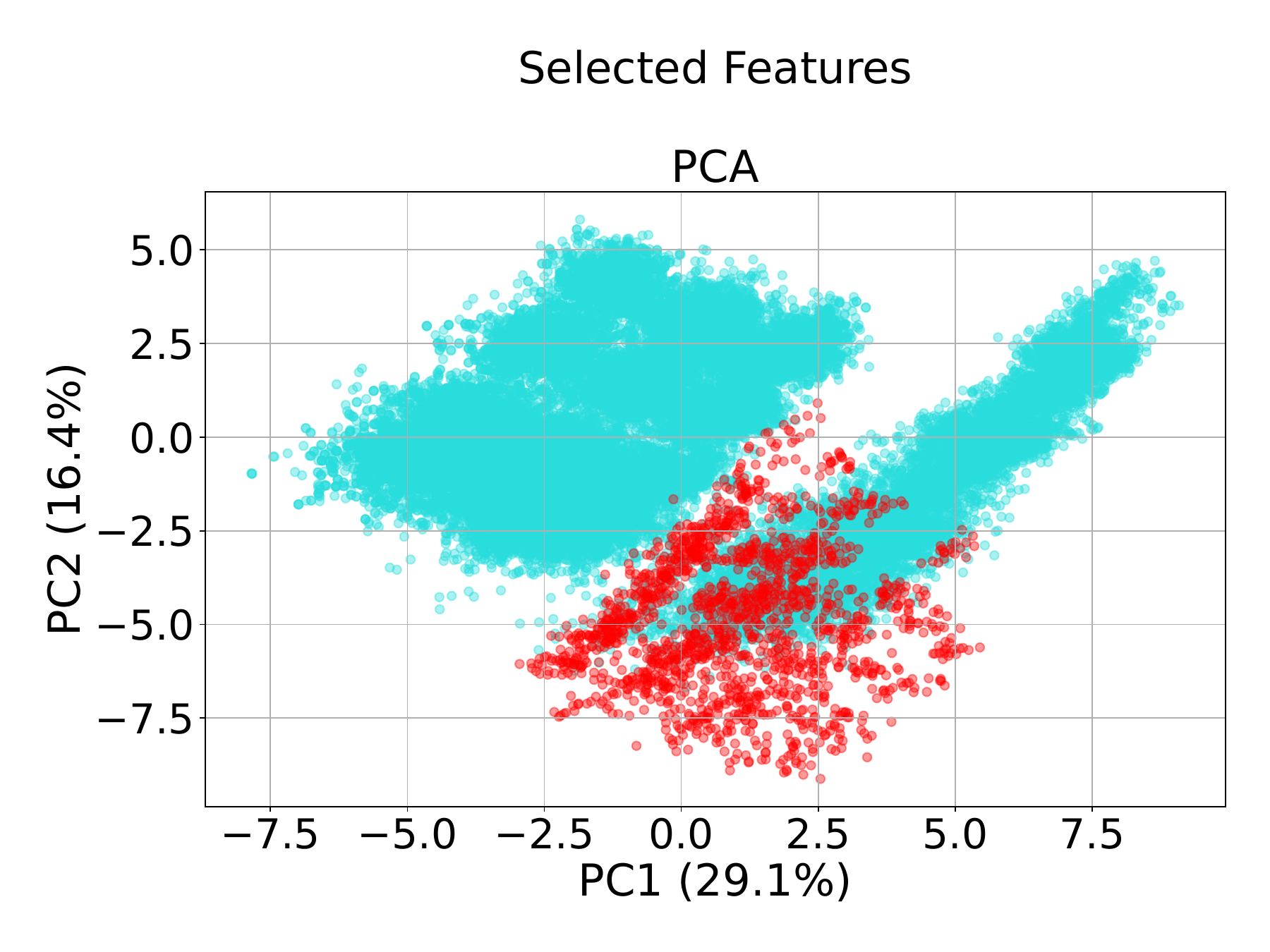}
    \includegraphics[width=0.49\linewidth]{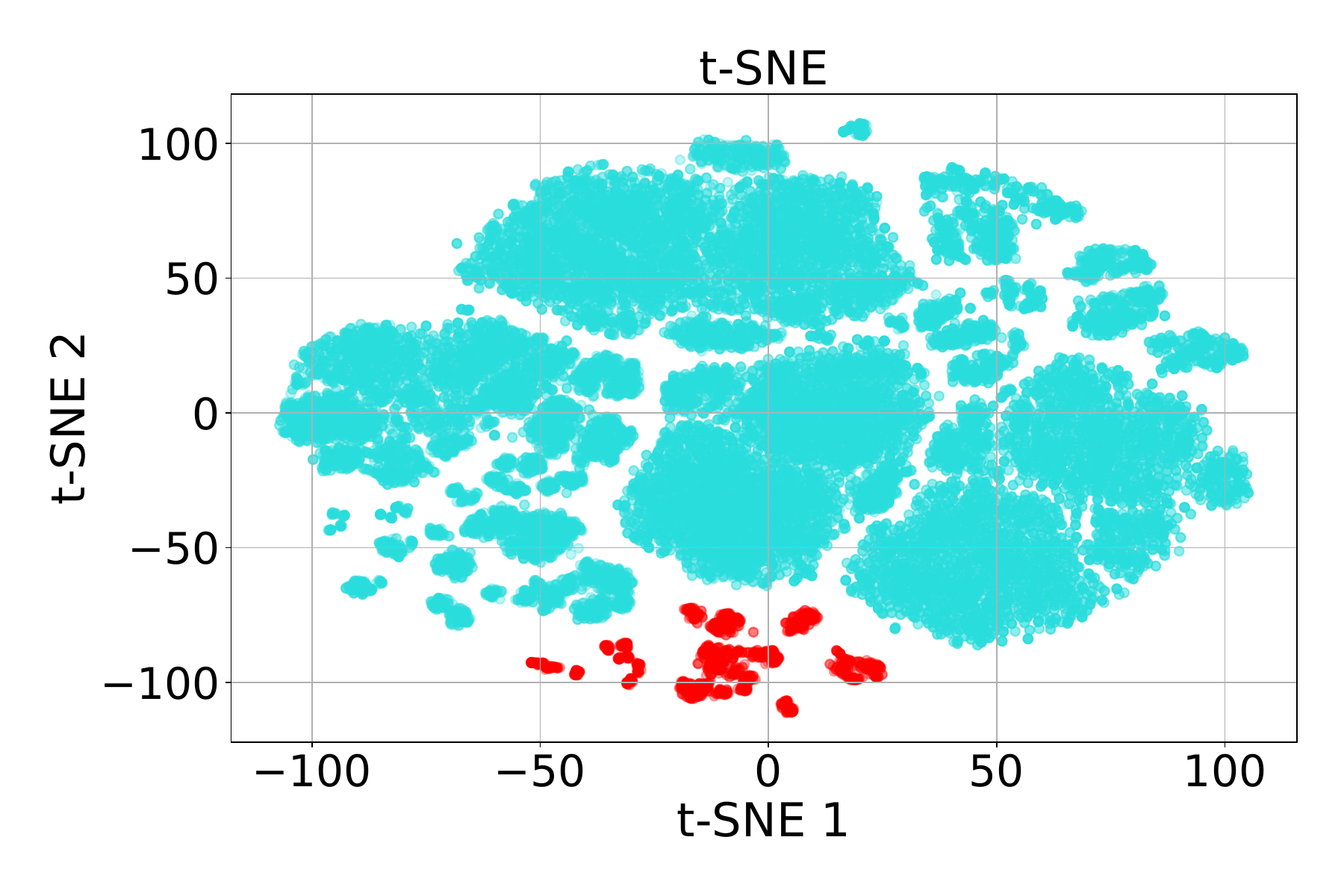}
    \includegraphics[width=0.49\linewidth]{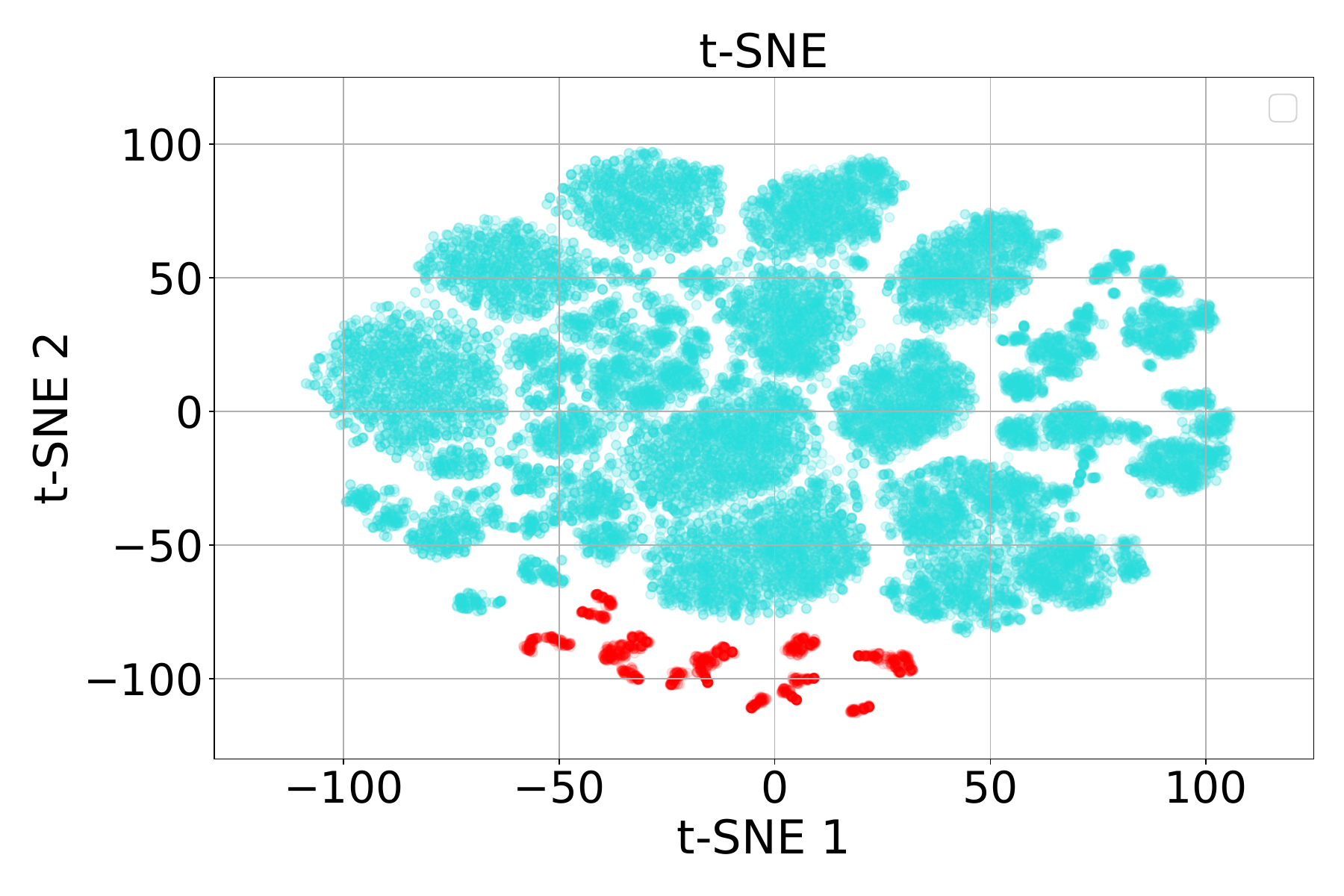}
    \caption{Dimensionality reduction using PCA and t-SNE for the tri-objective problems. The left plot uses all features. The right plot uses the selected features.}
    \label{fig:pca_by_benchmark_3obj}
\end{figure}

Figure \ref{fig:pca_by_benchmark_3obj} shows the scatter plots for the tri-objective problems.
The left column shows the plots of PCA and t-SNE based on all features. With t-SNE a clear separation between TriObjMPM2 and TriObjtDTLZ is visible while the plot based on PCA shows a small overlapt of the two benchmark sets. In the PCA plot four clusters can be seen within the benchmark sets can be seen. These clusters represent the different dimensionalities.
The plots based on the $38$ selected features (right column) look similar to the ones based on all features. For PCA you can still make out four clusters which represent the decision space dimensionality. However it has to be noted that TriObjtDTLZ spreads out a lot more without any structure. Part of the TriObjtDTLZ problems overlap with the TriObjMPM2 cluster containing the problems with decision space dimensionality $2$, indicating that these are problems with the same properties.
In the t-SNE plot you can see that the formed clusters of problems look more dense, but other than that you can not make out a distinct difference.

The visual analysis of the features shows that the feature selection reduces the noise in the data as well as reveals problems which share the same properties.

\begin{figure}
    \centering
    \includegraphics[width=0.4\linewidth]{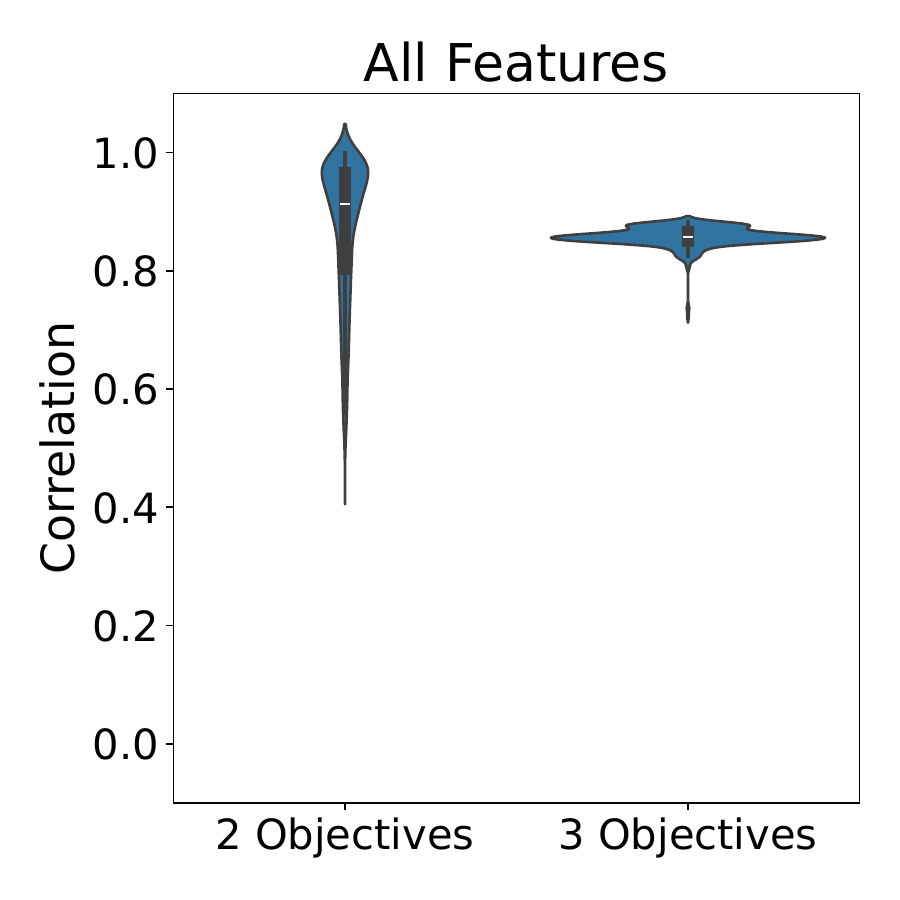}
    ~
    \includegraphics[width=0.4\linewidth]{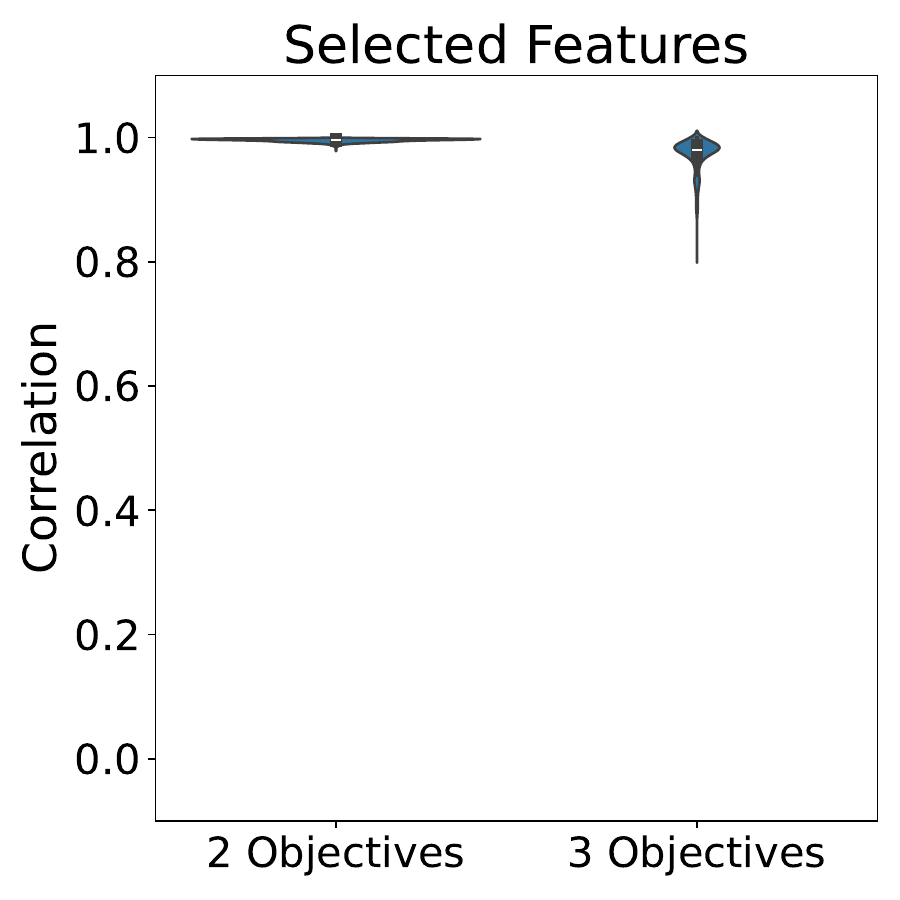}
    \caption{Correlations of the feature vectors for each individual problem.}
    \label{fig:problem_correlations}
\end{figure}

\paragraph{Feature Stability}
Another aspect we looked at regarding the features was how \emph{stable} they are across different samples for a specific function. For that, we took all feature vectors for a single function and calculated pairwise Pearson correlations between these vectors. For the pairwise feature calculation all $20$ samples of a function were used. This was repeated for each sample size. The features of a function across different dimensions were not considered.
Afterwards we calculated the mean correlation for each correlation calculation. That resulted in a total of $704$ bi-objective problems $\times$ $4$ sample sizes = $2816$ and $461$ tri-objective problems $\times$ $4$ sample sizes = $1844$ correlation values.
If we have an average high correlation it indicates that the calculated features are stable and thus agnostic to the random sample or the sample size. Figure \ref{fig:problem_correlations} shows a violin plot with these average correlations of all problems. For both bi- and tri-objective functions we have an average correlation of about $0.8$. This shows that the features are sufficiently stable. As in the previous analysis, we repeated this analysis with the reduced feature set chosen by the feature selection. The resulting violin plot (Fig. \ref{fig:problem_correlations}) shows an average correlation of close to $1$ without noteworthy outliers. This indicates that the selected features are almost perfectly stable and are thus not necessarily affected by the randomness the initial sample.

\paragraph{Feature Correlation}
Not only is the stability of the features important, but also their expressiveness. To measure that, we calculated the pairwise absolute Pearson correlations of the features themselves across all specific function, dimension and sample size combinations. Afterwards, we took the mean of these correlations. Ideally, we want to see a low correlation between features since that indicates that each feature describes different aspects of a problem. 

The average absolute correlation for the bi-objective problems is $0.30$ and for the tri-objective functions $0.23$. These low correlation values indicate that the selected features each describe different aspects of a given problem.

\begin{figure*}
    \centering
    \includegraphics[width=1\linewidth]{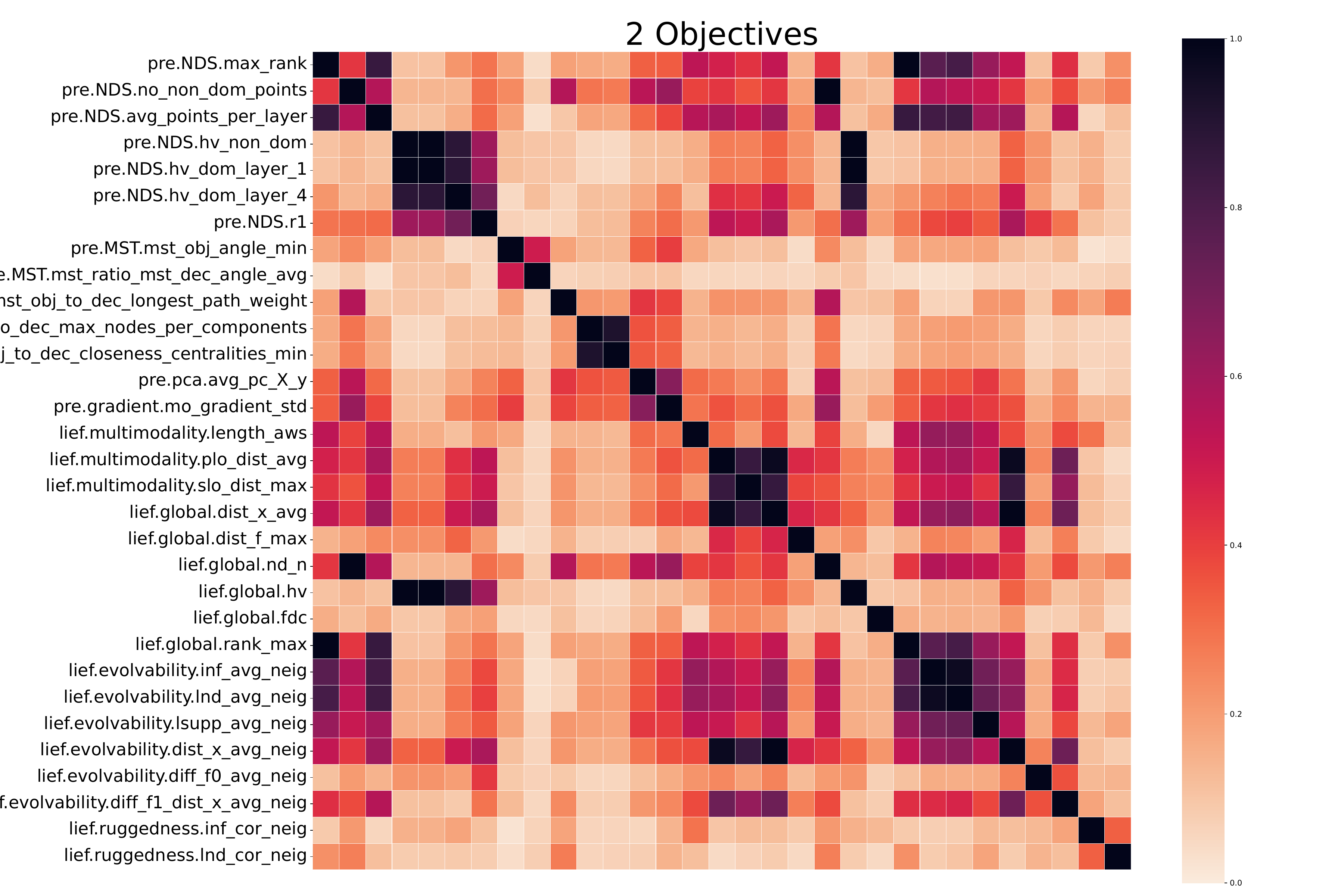}
    \includegraphics[width=1\linewidth]{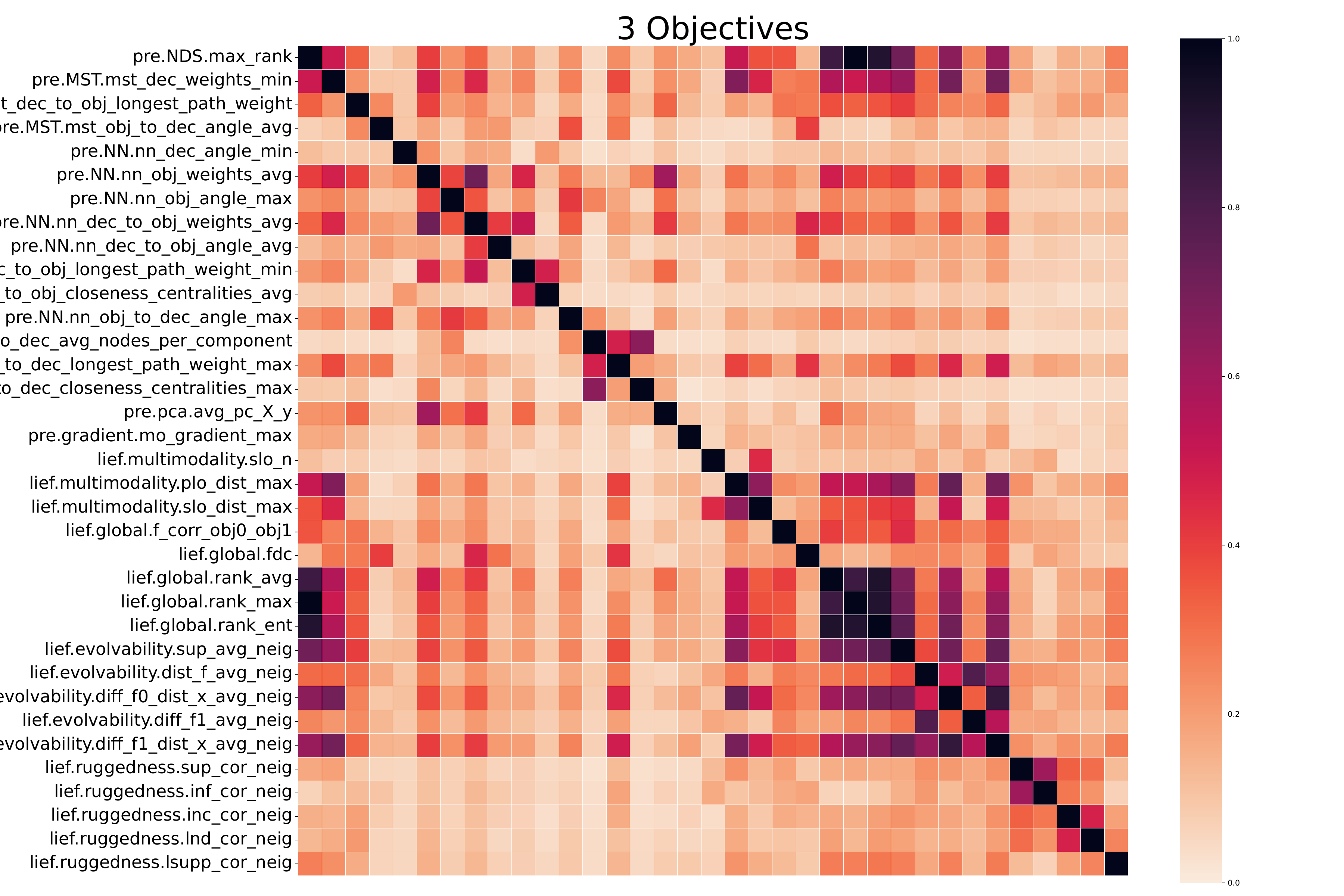}
    \caption{Absolute Pearson correlations of the selected features.}
    \label{fig:correlations_selected_feats}
\end{figure*}

\begin{figure}
    \centering
    \includegraphics[width=0.49\linewidth]{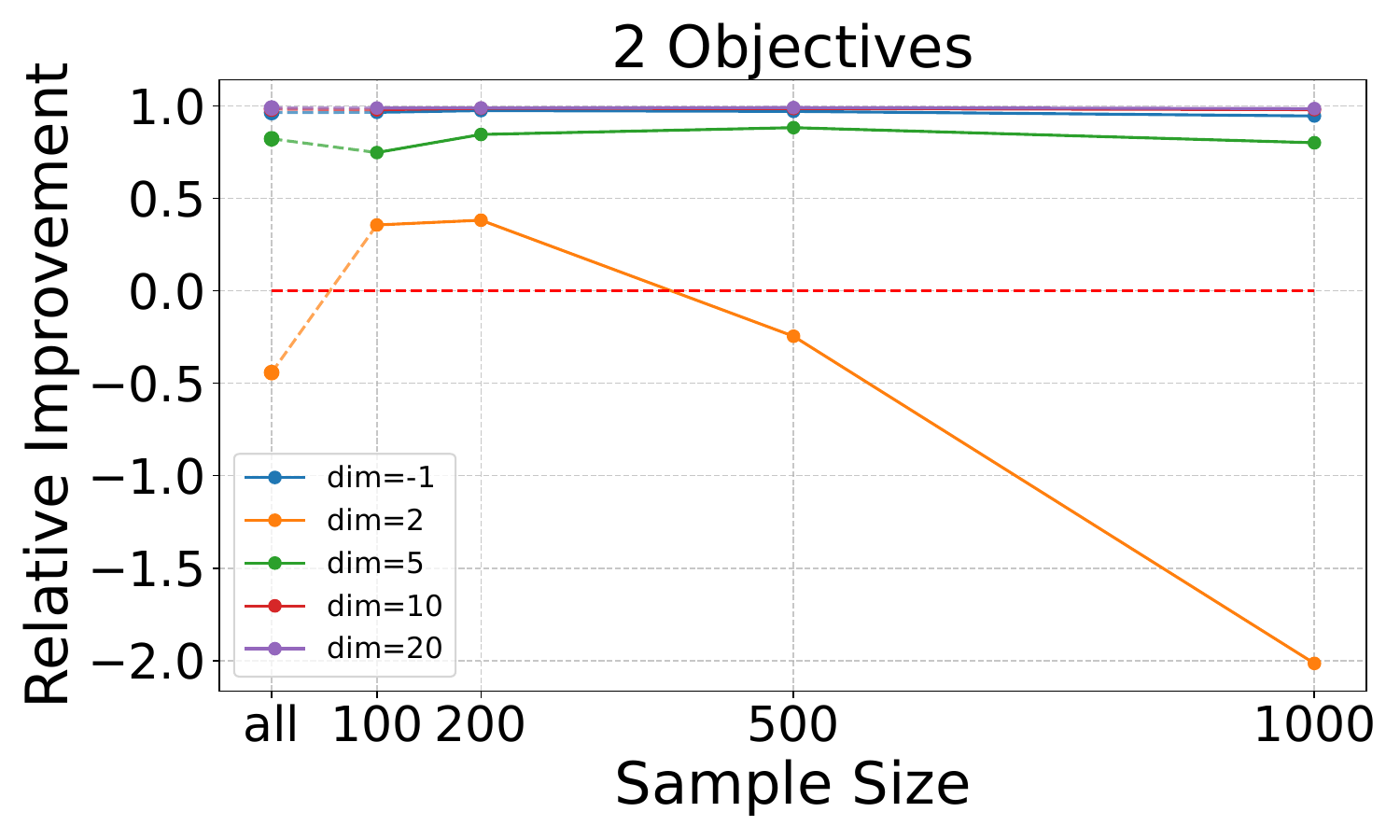}
    \includegraphics[width=0.49\linewidth]{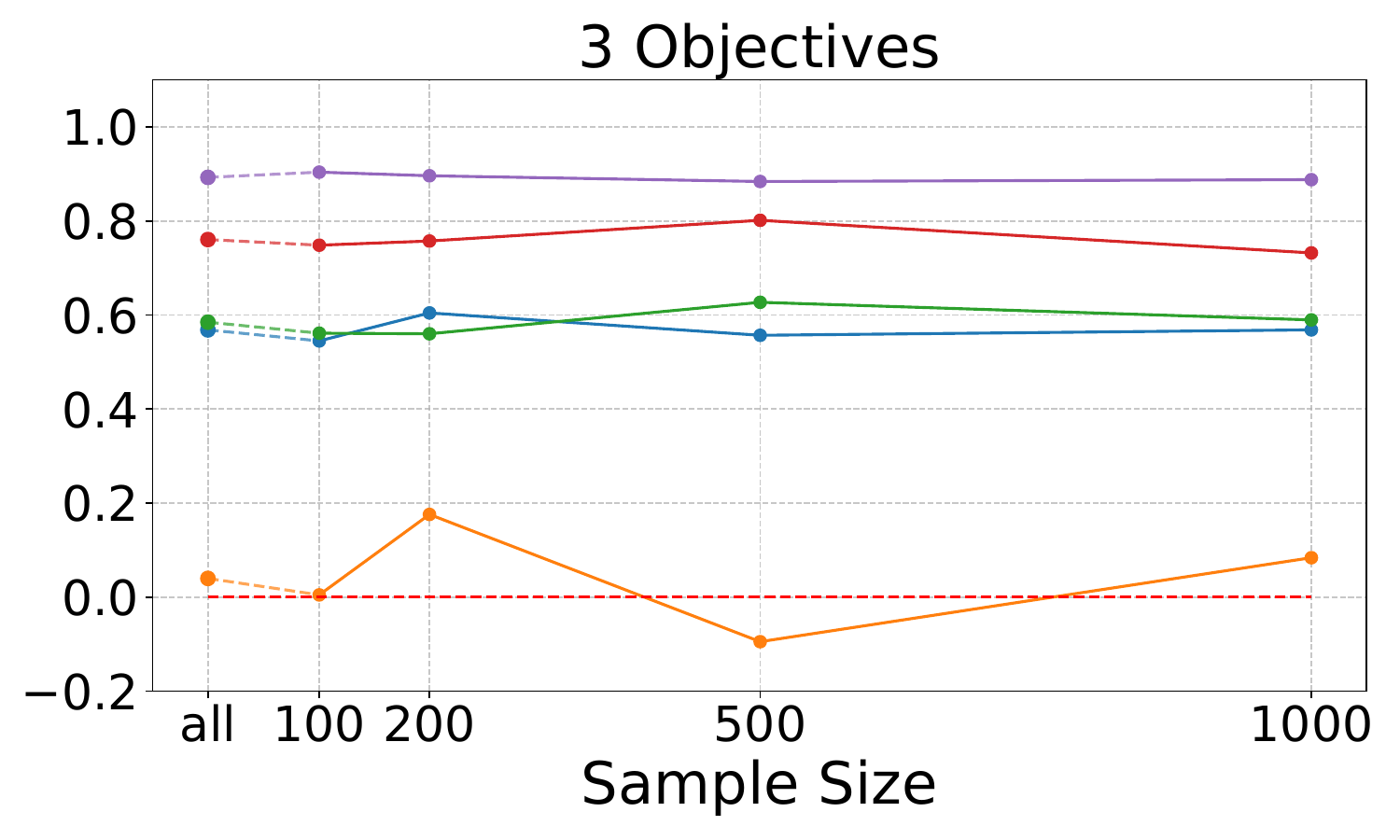}
    \caption{Relative improvement by dimension and sample size.}
    \label{fig:relative_improvements_by_dim_sample_size}
\end{figure}

\paragraph{Sample Size}
In the data set for the AAS, we use different sample sizes $s \in \{100, 200, 500, 1\,000\}$ for the feature calculation of the same functions with $20$ different seeds.
Here we want to investigate the difference in performance for the different sample sizes and dimensions. For that, we compared the relative improvement of the best selector on the test set for the different sample sizes and dimensions displayed in Figure \ref{fig:relative_improvements_by_dim_sample_size}. 
The sample size used for the feature calculation is subtracted from the budget used for the performance of the solvers. This ensures a fair comparison of the selector performance compared to the SBS.
First, we look at the bi-objective functions. 
If you look at dimension $10$ and $20$ you can see a constant high RI of above $0.98$. For dimension $10$ the highest RI is achieved by sample size $500$ with $0.993$ and the lowest RI is $0.983$ with sample size 100. Dimension $20$ has the highest RI of $0.994$ with sample size $500$ and the lowest RI of $0.987$ with sample size $1000$. It has to be observed here that the RI only differs at the third decimal place. This overall high RI can be explained by the overall budget of $15\,000$ and $25\,000$ (see Table \ref{tab:function_evals}) used for dimension $10$ and $20$ respectively. Considering the sample sizes used they make a small fraction of the overall budget which were subtracted resulting in a comparable high HVN for the solver runs. A more interesting structure can be observed for dimension $5$. Here the smallest RI of $0.75$ is achieved with a sample size of $100$ while the highest RI is achieved by sample size $500$ with a RI of $0.89$. You can clearly see that the RI improves by with a higher sample size with its peak at a sample size of $500$ and after that decreases with a sample size of $1000$. This shows that having a larger budget with a small sample size does not compensate for limited feature expressiveness. In contrast, increasing the sample size, even with a reduced budget, improves the RI. This indicates that a larger sample size, which leads to a better selector, is more important than the budget used. Dimension $2$ has the most interesting structure. While the highest RI of $0.47$ is achieved by a sample size of $200$ the smallest RI is achieved using sample size 1000 with a RI of $-2.35$. This means using a sample size of $1000$ decreases the performance of the solvers over the SBS drastically. The same hold for sample size $500$ with a RI of $-0.31$. This is explained by the small budget of $3000$ (see Table \ref{tab:function_evals}). Reducing the budget by $500$ or $1000$ decreases the solver performance by an amount that using a selector is not feasible any more.
These results show that for dimensions $5$, $10$ and $20$ a sample size of $500$ yields the best result while a sample size of $200$ still achieves comparable results. For dimension $2$ a sample size of $200$ yields the best result closely followed by sample size $100$.

Next, we look at the tri-objective functions. Here we can observe a similar pattern of the change in RI for the different dimension although the RI is overall worse when compared to the bi-objective functions. Dimension $20$ has the overall highest RI of $0.903$ with sample size $100$. The smallest RI of $0.883$ for dimension $20$ is reached by sample size $500$. Dimensions $5$ and $10$ follow the same structure. With increasing sample size the RI increases to $0.626$ and $0.801$ for dimension $5$ and $10$ respectively until it drops down for sample size $1000$. We saw this pattern with the bi-objective functions. With a larger sample size the expressiveness of the features increases and thus the selector can make better decisions. Even the smaller budget for the solvers does that yield a higher RI until sample size $1000$ where the reduced budget has an impact. The overall smallest RI has dimension $2$. Here the highest RI of $0.175$ is achieved with a sample size of $200$ while the smallest RI of $-0.094$ is achieved by sample size $500$. Here holds the same reasoning as for the bi-objective functions with dimension $2$. The budget of the solvers is reduced by an amount where picking the correct solver does not compensate for the reduced function evaluations. An outlier here is sample size $1000$ which has a larger RI, even positive, than sample size $500$.

\section{Discussion}
\label{sec:discussion}

Through the performed AAS we were able to show that our newly created features, in combination with existing features, are capable of distinguishing between the algorithm performance on various problems.
Especially for the bi-objective problems we almost closed the SBS / VBS gap completely, i.e. reached a RI close to $1$ resulting in always choosing the VBS to solve a problem. For the tri-objective problems, we achieved a RI of about $0.272$. 
Additionally, we showed that a small subset of all features is expressive enough to capture the distinguishing properties of the given problems with regarding to problem hardness related to the considering algorithms. The subset of features is agnostic to the chosen sample and sample size which makes them stable. 
A limitation of the AAS study is the use of HV as a singular performance metric. We will investigate other performance metrics in future studies.

An important advantage of the used features is the expressiveness with relatively small sample sizes. For the bi-objective functions, a sample size starting at $200$ is already sufficient to achieve a high RI. While our recommendation for the best RI is a sample size of $500$ for dimensions $5$, $10$ and $20$. For dimension $2$ the recommended sample size is $200$. Though, in settings with a very restrictive budget a sample size of $200$ or even $100$ is sufficient to achieve a high RI for all dimensions. For the tri-objective functions, the same sample sizes as for the bi-objective functions hold. For dimension $5$, $10$ and $20$ a sample size of $500$ achieves the best performance while the sample sizes $100$ and $200$ achieve sufficient RI ins restricted settings.
Connected to the discussion of the sample size is the runtime for feature calculation. Though explicit runtimes for the feature calculations will not be presented. In the used setting for sample sizes of $100$, $200$, $500$ and $1000$ the feature calculations were reasonably fast (i.e., seconds), with a strong relation between sample size and computation time. Furthermore the dimensionality of the decision and objective spaces also have an effect on the feature computation time.   
This results in the fastest feature computation for bi-objective problems with dimension $2$ and a sample size of $100$ while the longest feature computation results from tri-objective functions with dimension $20$ and a sample size of $1000$. 
When looking at the different feature groups we can observe different runtimes. The feature groups NDS, PCA, PCA, Gradient based and Descriptive can be computed relatively fast as their complexity scales linearly with the sample size in complexity. Graph-based feature groups, MST and NN, are more computationally expensive, as graph construction already has quadratic complexity, followed by the feature computation itself. NN-based features are particularly slow when many individual graphs are involved. Thus, in a setting where runtime is a critical resource the feature groups MST and NN could be dropped. Although these features can have a high importance as we can see with the tri-objective functions. Here, the feature selection resulted in a total of $36$ features of which $14$ were of the feature group MST and NN. A practitioner has to choose between the trade-off of higher expressiveness or faster runtime. 
However, the feature computation has to be seen relative to the time it takes to evaluate the target function. In general artificial benchmark sets are very fast which causes the feature computation to have a high fraction of the total runtime. Whereas real world problems or simulations can take longer to evaluate where in turn the feature computation becomes a small fraction of the total runtime.

An interesting aspect we observed was a visual separation of the benchmark sets while having an overlap of functions that share the same properties.
We further investigated correlations of the features from different aspects. The first aspect demonstrated that the features are stable with respect to the random sample. The second aspect investigated the correlation of the features with each other. Here we demonstrated that the features are mostly uncorrelated.

\section{Conclusion \& Future Work}
\label{sec:conclusion}

We introduced a novel set of exploratory landscape features tailored to box-constrained continuous multi-objective optimisation problems. By leveraging dominance relations, graph-based representations in both decision and objective spaces, gradient information and dimensionality reduction techniques our proposed features provide meaningful insights into problem structure. Empirical results on established benchmarks show that these features significantly enhance automated algorithm selection, effectively predicting algorithm performance and closely approximating the virtual best solver. 
Moreover, the selected features  are stable, expressive, and informative even with small sample sizes, making them practical for real-world applications.

The research regarding MO-ELA is still in its early steps. Thus, there are a lot of different things that can be looked at in further research. A first aspect to look at are different sample sizes as we showed that a small sample is sufficient to capture enough of a problem to perform AAS. 
This includes, among other things, investigating how the sample size affects different feature groups and how the runtime scales based on the dimensionality and the number of objectives.
Especially for the tri-objective problems a larger and more diverse benchmark set is needed to fully investigate the potential of the features. In addition to a larger and diverse benchmark set problems of higher dimensions in the decision space and potentially the objective space needs to be looked at. 
Extending the use-case of the features to other domains like automated algorithm configuration (AAC) can be promising. More specifically we investigated the performance of solvers configured by MO-SMAC~(~\cite{rook2025mo}), a multi-objective version of SMAC, with conflicting goals simultaneously~(\cite{preuss2024potential}). Here we see two potential applications of the features. First, use the features as additional information for the internal surrogate model of MO-SMAC. Second, investigate the landscape formed by the multi-objective configuration scenario itself. Another use case for the features might be per-instance  configuration as we now can describe the landscape of each instance individually.
For single objective ELA, it is possible to infer from low-level features to high-level properties. This would add a lot of value for MO-ELA. 
Currently it is important to scale the features in the decision and objective space to compare problems with arbitrary ranges with each other. Adapting the features to be agnostic to arbitrary ranges as well as the order of objectives and decision space variables is a crucial step to have robust features. Finally, comparing the MO-ELA features to feature-free deep learning approaches as it has been done with single objective ELA can prove insightful.

\small

\bibliographystyle{apalike}
\bibliography{bibfile}

\end{document}